\documentclass[10pt,twocolumn,letterpaper]{article}

\usepackage{cvpr}              % CAMERA-READY version
\definecolor{cvprblue}{rgb}{0.21,0.49,0.74}
\usepackage[pagebackref,breaklinks,colorlinks,allcolors=cvprblue]{hyperref}

\usepackage{enumitem, amsmath, booktabs, siunitx, multirow, arydshln, xcolor, float, colortbl, placeins}
\def\paperID{37495} 
\def\confName{CVPR}
\def\confYear{2026}

\usepackage[textsize=tiny]{todonotes}

\title{Rethinking Camera Choice: An Empirical Study on Fisheye Camera Properties in Robotic Manipulation}

\author{
Han Xue$^{1*}$\quad Nan Min$^{2*}$\quad Xiaotong Liu$^{3,4*}$\\ Wendi Chen$^{1,4}$\quad Yuan Fang$^{1,4}$\quad Jun Lv$^{5}$\quad Cewu Lu$^{1,4,5, \ddagger}$\quad Chuan Wen$^{1\ddagger}$\\
$^{1}$Shanghai Jiao Tong University \quad $^{2}$Southeast University\quad $^{3}$USTC \\
$^{4}$Shanghai Innovation Institute \quad $^{5}$Noematrix Ltd. \\
$^*$ Equal contribution.\quad $\ddagger$ Corresponding authors.
}

\begin{document}
\maketitle

\begin{abstract}
The adoption of fisheye cameras in robotic manipulation, driven by their exceptionally wide Field of View (FoV), is rapidly outpacing a systematic understanding of their downstream effects on policy learning. This paper presents the first comprehensive empirical study to bridge this gap, rigorously analyzing the properties of wrist-mounted fisheye cameras for imitation learning. Through extensive experiments in both simulation and the real world, we investigate three critical research questions: spatial localization, scene generalization, and hardware generalization. Our investigation reveals that: (1) The wide FoV significantly enhances spatial localization, but this benefit is critically contingent on the visual complexity of the environment. (2) Fisheye-trained policies, while prone to overfitting in simple scenes, unlock superior scene generalization when trained with sufficient environmental diversity. (3) While naive cross-camera transfer leads to failures, we identify the root cause as scale overfitting and demonstrate that hardware generalization performance can be improved with a simple Random Scale Augmentation (RSA) strategy. Collectively, our findings provide concrete, actionable guidance for the large-scale collection and effective use of fisheye datasets in robotic learning. More results and videos are available on \url{https://robo-fisheye.github.io/}.
\end{abstract}    
\section{Introduction}
Effective visual perception is a cornerstone of robust robotic manipulation. While standard pinhole cameras have been the default, there is a growing trend in robotics (\eg, UMI~\cite{chi2024universal}, RDT2~\cite{rdt2}, GEN-0~\cite{generalist2025gen0}, $\pi_{0.5}$~\cite{pi_0.5}) toward adopting fisheye cameras, often mounted on the robot's wrist. Characterized by an exceptionally large Field of View—often exceeding 180°—these cameras capture a significantly wider perspective than conventional counterparts. This wide-angle capability has already proven invaluable in domains like autonomous driving~\cite{9363560,9626604} and SLAM~\cite{jiPanoramicSLAMMultiple2020,10610351}, where comprehensive scene awareness is critical.

This emerging trend suggests a future reliant on large-scale, fisheye-based datasets for training imitation learning policies and Vision-Language-Action (VLA) models~\cite{black2024pi_0, pi_0.5, ghosh2024octo, generalist2025gen0, rdt2}. However, this adoption is rapidly outpacing our systematic understanding. The specific benefits and potential challenges of using wrist-mounted fisheye cameras specifically for robotic imitation learning remain largely unexplored. In this paper, we bridge this critical gap by conducting a comprehensive systematic analysis of the impact of fisheye camera characteristics on policy performance.

To provide actionable insights, we structure our analysis by examining the two defining characteristics of fisheye lenses and their direct impact on policy learning:
\begin{itemize}
    \item \textbf{(1) The Benefit of Wide FoV:} The primary advantage is the massive field of view. This naturally leads us to investigate its downstream effects on policy capabilities—specifically, how this enriched context improves a policy's understanding of the world.
    \item \textbf{(2) The Challenge of Distortion:} This wide FoV is achieved via severe radial distortion, a property absent in pinhole models. This presents a unique challenge, particularly for model generalization across different hardware with different intrinsic parameters.
\end{itemize}

Based on this framework, we formulate three key research questions. The first two are designed to probe the effects of the wide FoV's benefits, while the third directly confronts the challenge of hardware-specific distortion:
\begin{enumerate}
    \item \textbf{Spatial Localization:} To what extent does the wide FoV enhance a policy's spatial reasoning and localization capabilities?
    \item \textbf{Scene Generalization:} Do fisheye cameras improve a policy's robustness and generalization ability against novel or distracting backgrounds?
    \item \textbf{Hardware Generalization:} How well do policies trained on one fisheye camera transfer to a new, unseen fisheye lens with different intrinsic parameters?
\end{enumerate}

To rigorously address these questions, we conduct extensive experiments in \textit{both} simulation and the real world. For our simulation experiments, we implement a realistic fisheye camera model within the MuJoCo~\cite{todorov2012mujoco} physics engine. This allows us to conduct large-scale experiments across multiple tasks from two widely accepted imitation learning benchmarks, Robomimic~\cite{robomimic} and MimicGen~\cite{mandlekar2023mimicgen}, utilizing diverse combinations of camera and background settings. For our real-world validation, we employ multiple distinct physical cameras and backgrounds across three manipulation tasks and design a rigorous evaluation protocol to verify our findings from simulation.

Our exhaustive investigation uncovers several key findings and provides actionable guidance for scaling up robot data collection with fisheye cameras in the real world:
\begin{enumerate}
    \item \textbf{Fisheye cameras enhance spatial localization, but this is dependent on intra-scene complexity.} We find that the wider FoV captures more environmental feature points and enhances policy's spatial localization. However, this advantage diminishes in environments lacking distinct visual features (\eg, solid-colored backgrounds).
    
    \textbf{Guidance}: \textit{Prioritize data collection in visually complex and feature-rich environments.}

    \item \textbf{Fisheye cameras require inter-scene diversity to prevent overfitting.} While fisheye-trained policies can overfit more easily to simple scenes, their scene generalization capability surpasses that of standard cameras if trained with sufficient diversity in backgrounds (\eg textures, lighting, and distractors).
    
    \textbf{Guidance}: \textit{Maximize environmental diversity during data collection to unlock the generalization potential.}

    \item \textbf{Cross-camera generalization for fisheye lenses is a notable challenge, but one that is partially addressable.} We observe that naively transferring a policy to a new fisheye lens can cause a sharp performance drop. However, we demonstrate that our proposed Random Scale Augmentation (RSA) are effective at mitigating this performance drop. This suggests that the challenge is not insurmountable and that data-centric approaches are promising directions.
    
    \textbf{Guidance}: \textit{Use strong scale-oriented data augmentation during training to improve cross-camera transfer capability.}
\end{enumerate}
Taken together, our findings and guidance provide a rigorous foundation for the robotics community, enabling the confident and effective large-scale adoption of fisheye cameras for training the next generation of generalist policies.
\section{Related Work}

\noindent \textbf{Imitation learning for Robotic Manipulation.}
Imitation learning is a powerful paradigm for teaching robots complex manipulation skills from expert demonstrations. At its core lies Behavioral Cloning (BC), which formulates policy learning as a supervised problem of mapping observations to actions~\cite{10.5555/2969735.2969771,Schaal1999IsIL,5480475}. Recently, the field has undergone a paradigm shift toward diffusion-based policies~\cite{chi2024diffusionpolicy,ze2024d,pmlr-v270-fu25b,ha2024scaling,chi2024universal,huDataScalingLaws2024,pmlr-v270-ke25a}, which excel at modeling complex, multi-modal action distributions.
Despite these advances, imitation learning still faces challenges such as distributional shift~\cite{dagger2011,gail2016} and partial observability~\cite{de2019causal,wen2020fighting}. A key strategy to mitigate these issues is to collect abundant, high-quality demonstrations~\cite{pi_0.5,generalist2025gen0,black2024pi_0,huDataScalingLaws2024}, encompassing diverse behavioral trajectories and sensor observations with rich contextual information and large receptive fields.
In this work, we show that fisheye cameras, with their wide field of view (FoV), substantially enhance the contextual information available to the policy compared to standard pinhole cameras. We further conduct a systematic analysis of their impact, demonstrating performance gains and improved scene and hardware generalization enabled by the unique characteristics of fisheye cameras.

\noindent \textbf{Fisheye Cameras for Robotics.}
 Fisheye cameras are widely utilized for their large FoV in fields such as autonomous driving~\cite{9363560,9626604,9636707,9340732}, SLAM~\cite{jiPanoramicSLAMMultiple2020,2018arXiv181112633W,10610351}, surveillance~\cite{yangPanoramicUAVSurveillance2019,konradOverheadFisheyeCameras2024}, and aerial robotics~\cite{hausbergRelativeDroneGroundVehicle2020,AutonomousAerialRobot}. Recently, their adoption has extended to robotic manipulation to broaden workspace observation~\cite{chi2024universal,generalist2025gen0,huDataScalingLaws2024,wang2024dexcap}. Among these, UMI~\cite{chi2024universal} utilized fisheye cameras for portable data collection, providing valuable application-level insights. However, a systematic analysis isolating the quantitative impact of specific optical properties on policy learning remains absent. This gap is reflected in the current benchmarking landscape: existing fisheye datasets~\cite{yogamaniWoodScapeMultiTaskMultiCamera2019,9814840,Scheck_2020_WACV} lack robotic manipulation tasks, while popular robotics benchmarks ~\cite{robomimic,jiangDexMimicGenAutomatedData2025,9001253,2021arXiv210714483M} omit fisheye streams. Our work aims to fill this void by presenting a systematic study of fisheye cameras in robotic manipulation through both simulated and real-world experiments, culminating in a set of practical guidelines for their effective deployment.

\noindent \textbf{Fisheye Simulation.}
Fisheye cameras are widely used in real-world robotic manipulation, yet standard simulators lack fisheye rendering, hindering benchmarking and limiting the study of fisheye-based policies. To support consistent experimentation across real and simulated settings, it is therefore important to enable reliable fisheye simulation. Several approaches can synthesize fisheye images from pinhole renderings: 3D Gaussian Splatting~\cite{kerbl3Dgaussians} enables high-fidelity fisheye synthesis~\cite{liuEveryCameraEffect2025,Shin_2025_ICCV,DBLP:journals/corr/abs-2411-15355}, and diffusion models also show promise~\cite{fangCameraSettingsTokens2024,10.1007/978-3-031-72980-5_9}, but both remain too slow or resource-intensive for interactive simulation. In contrast, classical projection models~\cite{9170271} and toolkits such as OmniCV-Lib~\cite{Sadekar2020OmniCVLib} offer a stable and efficient alternative through a two-stage rendering pipeline that first produces a panoramic intermediate view and then reprojects it into the fisheye domain. To meet the requirements of stability, accuracy, and efficiency, we design a two-stage pipeline inspired by OmniCV-Lib~\cite{Sadekar2020OmniCVLib} for fisheye simulation.
\begin{figure*}[ht!]
  \centering
   \includegraphics[width=0.9\linewidth]{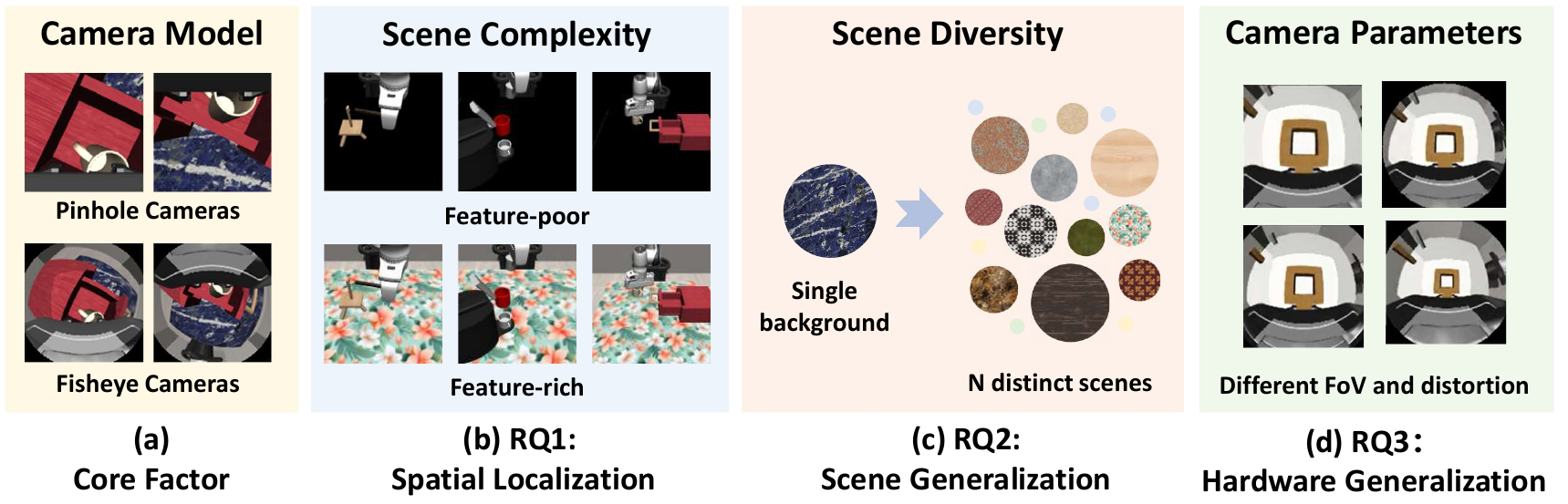}

   \caption{Overview of the four factors analyzed to address our Research Questions (RQs). We study:  \textbf{(a) Camera Model} (fisheye vs. pinhole) as our primary comparison; \textbf{(b) Scene Complexity} (poor vs. rich) for\textit{ spatial localization} (RQ1); \textbf{(c) Scene Diversity} (1 vs. N scenes) for \textit{scene generalization} (RQ2); and \textbf{(d) Camera Parameters} (varied intrinsics) for \textit{hardware generalization} (RQ3).}
   \label{fig:factors}
\end{figure*}

\section{Study Design}
\label{sec:study_design}
We design a comprehensive experimental framework spanning both simulation and the real world to perform our investigation of fisheye camera properties. This section details our problem formulation (\cref{subsec:formulation}), the core methodological components we developed to enable this study (\cref{subsec:enabling_components}), our imitation learning framework (\cref{subsec:il_framework}), our simulation (\cref{subsec:sim_setup}) and real-world (\cref{subsec:real_setup}) experimental setups, and our evaluation protocol (\cref{subsec:evaluation}).

\subsection{Problem Formulation}
\label{subsec:formulation}

Our study is designed to systematically investigate the impact of fisheye lenses on imitation learning. Our analysis is guided by three key questions that stem from the core properties (\eg wide Fov and distortion) of fisheye cameras:
\begin{enumerate}
    \item \textbf{(RQ1)} \textbf{Spatial Localization:} To what extent does the wide FoV of fisheye cameras enhance a policy's spatial localization capabilities?
    \item\textbf{(RQ2)} \textbf{Scene Generalization:} Do fisheye cameras improve a policy's robustness and generalization ability against novel or distracting backgrounds?
    \item\textbf{(RQ3)} \textbf{Hardware Generalization:} How well do policies trained on one fisheye camera transfer to an unseen fisheye lens with different intrinsic parameters?
\end{enumerate}

    To rigorously address these questions, we define a set of core factors for our analysis, as illustrated in \cref{fig:factors}. These factors serve as the independent variables in our experiments to isolate and measure their specific effects:
    \begin{itemize}
            \item \textbf{Camera Model:} This is our primary dimension of comparison. In all subsequent experiments across varying factors of analysis, we directly compare the performance characteristics of the \textit{Fisheye Camera} with those of a \textit{Standard Pinhole Camera} as a control group. Our specific focus is on \textit{wrist-mounted} cameras, a configuration adopted by recent large-scale in-the-wild data collection projects (\eg RDT2~\cite{rdt2}, GEN-0~\cite{generalist2025gen0}), where the short observation distance naturally amplifies the differences between the two camera models.

            \item \textbf{Scene Complexity:} To address RQ1 (Spatial Localization), we establish two control groups for the background settings: \textit{feature-poor} (\eg, solid color) and \textit{feature-rich} (\eg, textured). By comparing performance across these two extremes, we evaluate how the background texture availability affects the model's spatial localization ability.

            \item \textbf{Scene Diversity:} To address RQ2 (Scene Generalization), we establish a scaling methodology ranging from a single background to $N$ distinct scenes during training. The efficacy of this increased diversity is then rigorously tested by measuring the policy's zero-shot transfer performance when deployed in completely \textit{unseen scenes} that were not included in the training distribution.

            \item \textbf{Camera Parameters:} To address RQ3 (Hardware Generalization), we evaluate the policy's cross-hardware generalization by assessing its performance when deployed with \textit{unseen camera intrinsics} (i.e., different FoV and distortion profiles) in a zero-shot manner .
        \end{itemize}

\begin{figure}[ht!]
  \centering
   \includegraphics[width=0.8\linewidth]{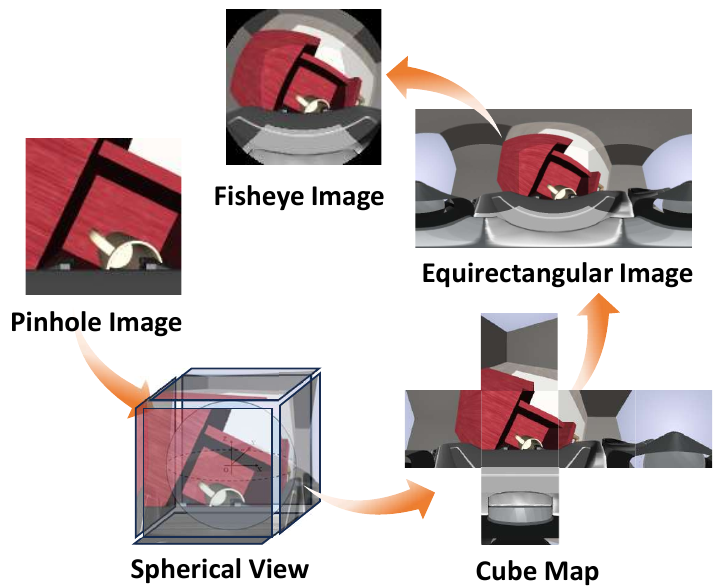}

   \caption{The implementation pipeline of fisheye camera simulation in MuJoCo~\cite{todorov2012mujoco}.}
   \label{fig:fisheye_sim}
\end{figure}

\subsection{Enabling Fisheye-centric Imitation Learning}
\label{subsec:enabling_components}
To investigate our research questions, particularly RQ1 and RQ3, we first had to address two significant technical gaps: the lack of realistic fisheye camera support in standard simulators and the challenge of cross-camera generalization. We introduce two components to solve these issues.

\begin{figure}[htbp!]
  \centering
  % \fbox{\rule{0pt}{1.5in} \rule{0.9\linewidth}{0pt}}
   \includegraphics[width=0.8\linewidth]{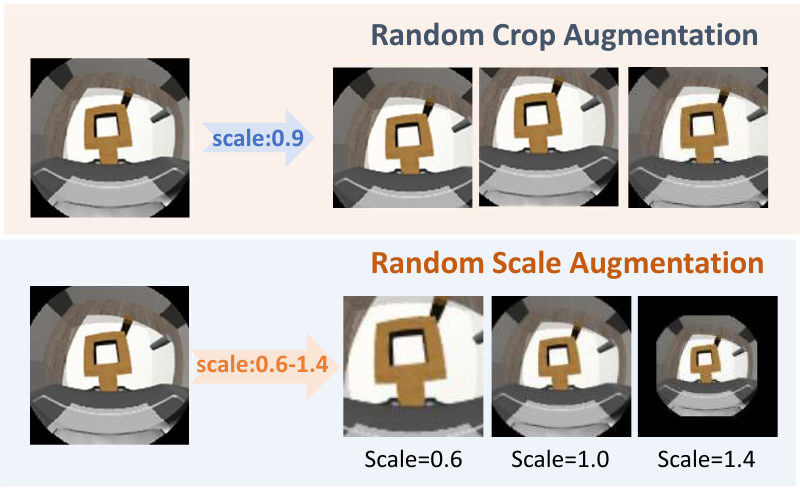}

   \caption{Random Crop Augmentation (fixed scale) vs. Random Scale Augmentation (RSA) for cross-camera generalization. }
   \label{fig:scale_augmentation}
\end{figure}

\textbf{Fisheye Camera Simulation in MuJoCo.}
To the best of our knowledge, no existing robotics simulation benchmark provides native support for fisheye cameras. We therefore implement a fisheye camera model in the MuJoCo~\cite{todorov2012mujoco} physics engine, adapted from the pipelines used in Robomimic~\cite{robomimic} and MimicGen~\cite{mandlekar2023mimicgen}. We employ a two-stage projection pipeline to simulate a fisheye camera, which is illustrated in \cref{fig:fisheye_sim}. The process begins by capturing a full 360-degree spherical view within our simulation environment. To achieve this, we place six virtual cameras, each oriented along a cardinal direction (front, back, left, right, up, and down), and assemble their six resulting images into a single \textit{cubemap}. In the first stage of the pipeline, the six faces of this cubemap are projected and stitched together to form an intermediate panoramic representation called an \textit{equirectangular image}. This common format effectively ``unwraps" the spherical view onto a 2D plane. In the second stage, this equirectangular image is transformed into the final fisheye view by applying a specific projection model, allowing us to simulate various lens characteristics. The entire process is implemented using functionalities for omnidirectional cameras inspired by libraries such as OmniCV-Lib~\cite{Sadekar2020OmniCVLib}.

\textbf{Random Scale Augmentation (RSA).}
To address RQ3 and mitigate the sharp performance drop observed during cross-camera generalization, we introduce a simple yet highly effective data augmentation technique we term Random Scale Augmentation (RSA), shown in \cref{fig:scale_augmentation}. We hypothesize that the primary challenge for policies transferred to unseen fisheye lenses is the significant variation in object scale induced by different lens intrinsics.  Our RSA method directly addresses this by forcing the policy to become robust to scale variations. Instead of a fixed-scale crop (\eg, 0.95), we sample a random scale factor $s$ from a wide uniform distribution (\eg, $U(0.7, 1.3)$) for each training image. The image is then center-cropped to this scale and resized to the standard network input size. Critically, if $s > 1.0$, this operation effectuates a ``zoom-out," where the source image is resized down and the surrounding canvas is padded with black. We posit that this simple but effective augmentation strategy prevents the network from overfitting to the absolute pixel scale of objects. Instead, it compels the policy to learn relative spatial relationships, such as the scale of the target object relative to the robot's end-effector, which is a more generalizable cue across different camera systems.

\subsection{Imitation Learning Framework}
\label{subsec:il_framework}
We build our system upon a standard, state-of-the-art visual imitation learning framework. Our design choices are detailed below, following a logical flow from the core algorithm to its inputs and outputs.

\textbf{Core Algorithm.} We employ the Diffusion Policy~\cite{chi2024diffusionpolicy} framework to model the extensive visual data collected. This choice is motivated by its demonstrated excellence in real-world manipulation tasks and its emergence as a powerful and widely accepted baseline for vision-based robotic imitation learning. Following standard practice, we utilize a U-Net~\cite{u-net} architecture as the noise prediction network, and we employ the DDIM~\cite{ddim} scheduler for efficient inference.

\textbf{Policy Inputs.} We design the policy to be \textit{state-free}, relying exclusively on visual data. This is a critical decision to directly test our RQs.
\begin{itemize}
    \item \textbf{Visual Encoder:} We specify different encoders tailored to the needs of each domain:
    \begin{itemize}
        \item \textbf{Simulation:} For controlled large-scale experiments in simulation, we utilize a standard ResNet-18~\cite{resnet} without pre-training. This serves as a widely adopted, robust, and computationally efficient baseline, ensuring that our results are comparable with existing simulation benchmarks. 
        \item \textbf{Real-World:} For real-world validation, where domain shift and visual fidelity are greater concerns, we leverage the features extracted by the CLIP~\cite{clip} Vision Transformer (ViT~\cite{vit}). We utilize its powerful pre-trained vision features to ensure enhanced robustness against novel textures and lighting conditions inherent in physical environments.
       
    \end{itemize}
 \item \textbf{No Proprioception Input:} A core design decision for this analysis is the \textit{omission of proprioceptive state} (\eg, end-effector pose and joint states) during policy training. This creates a state-free policy, which forces the model to rely exclusively on the visual input for spatial localization. As demonstrated in prior work on spatial generalization~\cite{state-free-policy}, removing proprioception prevents the policy from overfitting to simple state vectors and directly isolates and tests the visual encoder's capability—and specifically the fisheye camera's FoV advantage—in tasks that require strong visual cues for localization.
 
\textbf{Policy Outputs (Action Space).} For our simulation experiments, we adhere to the default settings of Robomimic~\cite{robomimic}, employing \textit{delta action} (relative transformations between consecutive frames) as the action space. For real-robot experiments, we adopted the default settings from UMI~\cite{chi2024universal}, adopting \textit{relative action} (relative transformations to the first frame of an action chunk). Prior work~\cite{state-free-policy} has shown that \textit{relative action space} provide superior spatial generalization compared to \textit{absolute action}, particularly in the absence of proprioceptive input.
\end{itemize}

\subsection{Simulation Experimental Setup}
\label{subsec:sim_setup}

\begin{itemize}
    \item \textbf{Benchmarks:} Two MuJoCo-based~\cite{todorov2012mujoco} benchmarks, Robomimic~\cite{robomimic} and MimicGen~\cite{mandlekar2023mimicgen}, which are adapted to enable fisheye camera rendering.

    \item \textbf{Camera Configurations:} We define two camera configurations for experiments in \cref{fig:fisheye_sim} (b):
        \begin{itemize}
            \item \textbf{Pinhole Camera:} One or two wrist-mounted pinhole cameras (90° FoV), no third-view camera.
            \item \textbf{Fisheye Camera:} One or two wrist-mounted fisheye cameras (235° FoV), no third-view camera.             
        \end{itemize}
        We deliberately exclude third-view cameras in our setup, consistent with prior UMI~\cite{chi2024universal}-like works (\eg RDT2~\cite{rdt2}, GEN-0~\cite{generalist2025gen0}). This exclusion is to avoid introducing confounding variables that would complicate an isolated analysis of the fisheye camera's effects.

    \item \textbf{Tasks:} We select six challenging tasks from Robomimc~\cite{robomimic} and MimicGen~\cite{mandlekar2023mimicgen} as shown in \cref{fig:sim_task}. These tasks assess capabilities across different dimensions, including: \textbf{high-precision manipulation} (\textit{Tool Hang}, \textit{Threading}), \textbf{spatial generalization} (\textit{Square}, \textit{Assembly}), and \textbf{long-horizon task execution} (\textit{Coffee}, \textit{Mug Cleanup}).
\end{itemize}

\begin{figure}[htbp!]
  \centering
  % \fbox{\rule{0pt}{1.5in} \rule{0.9\linewidth}{0pt}}
   \includegraphics[width=0.8\linewidth]{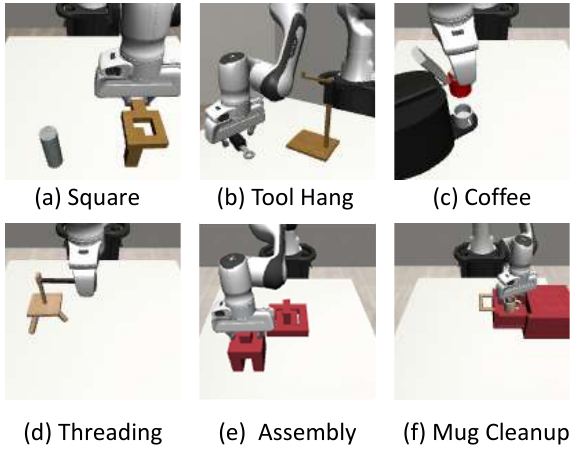}

   \caption{The six tasks in simulation experiments.}
   \label{fig:sim_task}
\end{figure}

\subsection{Real-World Experimental Setup}
\label{subsec:real_setup}

\begin{figure*}[htbp!]
  \centering
  % \fbox{\rule{0pt}{2in} \rule{\linewidth}{0pt}}
   \includegraphics[width=\linewidth]{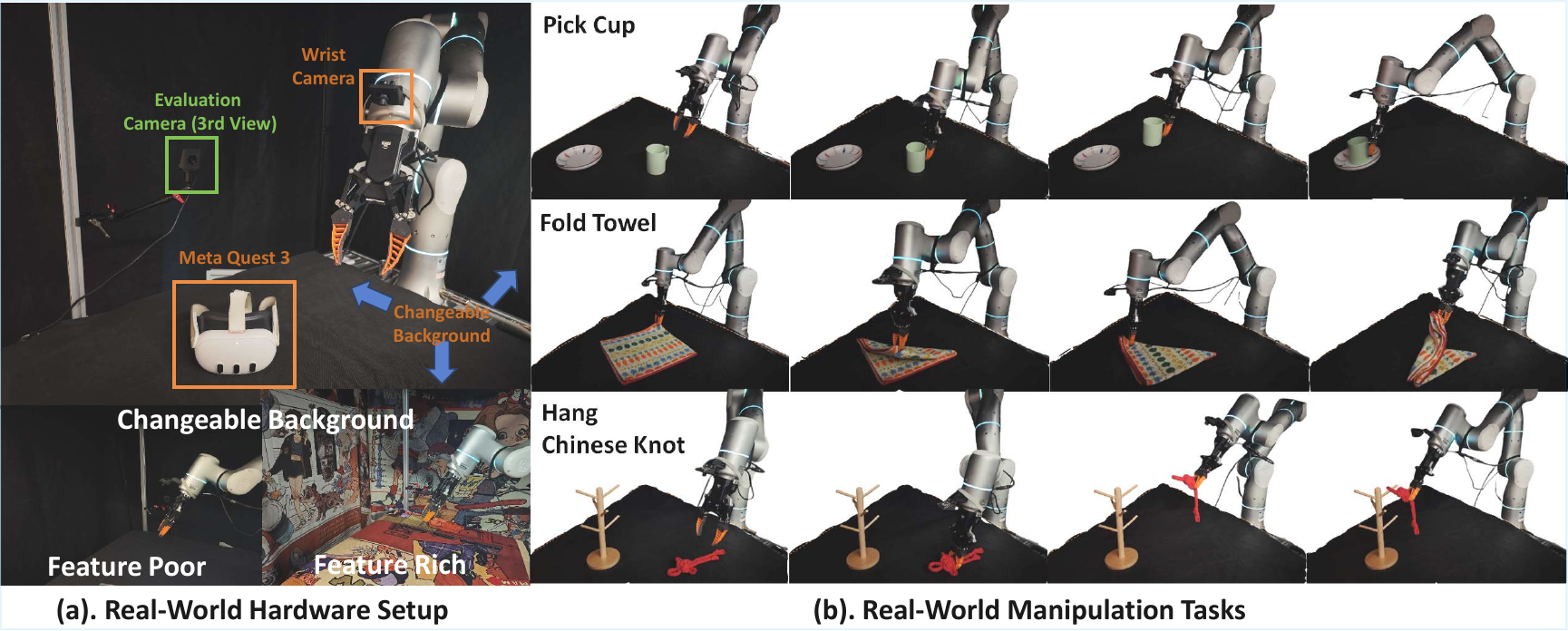}

   \caption{(a) The real-world experiment setup, which includes \textbf{changeable backgrounds} for \textit{scene complexity} (RQ1) and \textit{scene generalization} (RQ2) experiments. (b) The three tasks in real-world experiments: \textit{Pick Cup}, \textit{Fold Towel} and \textit{Hang Chinese Knot}.}
   \label{fig:realworld_setup}
\end{figure*}

\begin{itemize}
    \item \textbf{Hardware Platform:} We use one Flexiv Rizon 4~\cite{FlexivRizon} robot arm and DH AG-160-95 gripper~\cite{DHRoboticsAG} as the hardware platform, as shown in \cref{fig:realworld_setup}. The data collection is performed by teleoperation with Meta Quest 3~\cite{MetaQuest3}.
    \item \textbf{Camera Configuration:} We use two camera configurations (\cref{fig:realworld_setup}) without third-view cameras which aligns with the simulation setting:
        \begin{itemize}
            \item \textbf{Pinhole Camera:} One wrist-mounted pinhole camera (60° FoV), no third-view camera.
            \item \textbf{Fisheye Camera:} One wrist mounted fisheye camera (180° FoV), no third-view camera.
        \end{itemize}

    \item \textbf{Tasks:} We design three tasks as shown in \cref{fig:realworld_setup}, which test various robotic manipulation skills. These include \textbf{spatial generalization} (\textit{Pick Cup}), \textbf{deformable object manipulation} (\textit{Fold Towel}), and \textbf{high-precision rotational manipulation} (\textit{Hang Chinese Knot}).
\end{itemize}

\subsection{Evaluation Protocol}
\label{subsec:evaluation}

We conduct rigorous evaluations in both simulation and the real world to assess policy performance and generalization. Our protocols are designed to ensure reliability and fair comparisons across different camera systems.
\begin{itemize}
    \item \textbf{Simulation Experiments}
    For our simulation-based analysis, we adhere to the standard evaluation pipelines established by the Robomimic and MimicGen benchmarks. Performance is quantified using the \textbf{Success Rate (SR)}. For each experimental run, we select the policy checkpoint with the highest performance during training. We then execute this policy for 50  rollouts for evaluation.

\item \textbf{Real-World Experiments}
Our real-world protocol is designed to provide a granular and robust measure of performance while ensuring fairness. In complex real-world manipulation, a binary success rate is often too sparse to capture nuanced policy behaviors. Following prior work \cite{huDataScalingLaws2024}, we define a normalized, multi-stage scoring metric. Each task is decomposed into several key stages (typically 2--3). A policy receives a point for successfully completing each stage, and we report a final \textbf{Normalized Score}:
$$
\text{Normalized Score} = \frac{\text{Total points earned}}{\text{Total number of stages}}
$$
This metric, averaged over $N=20$ trials for each setup, provides a far more granular signal of policy capability than a simple pass/fail metric. To enable a fair and direct comparison across all experimental conditions, we meticulously control the experimental setup. Before every single rollout, we reset the robot pose and all relevant object poses to pre-defined initializations.

\end{itemize}

\begin{figure}[ht!]
  \centering
   \includegraphics[width=\linewidth]{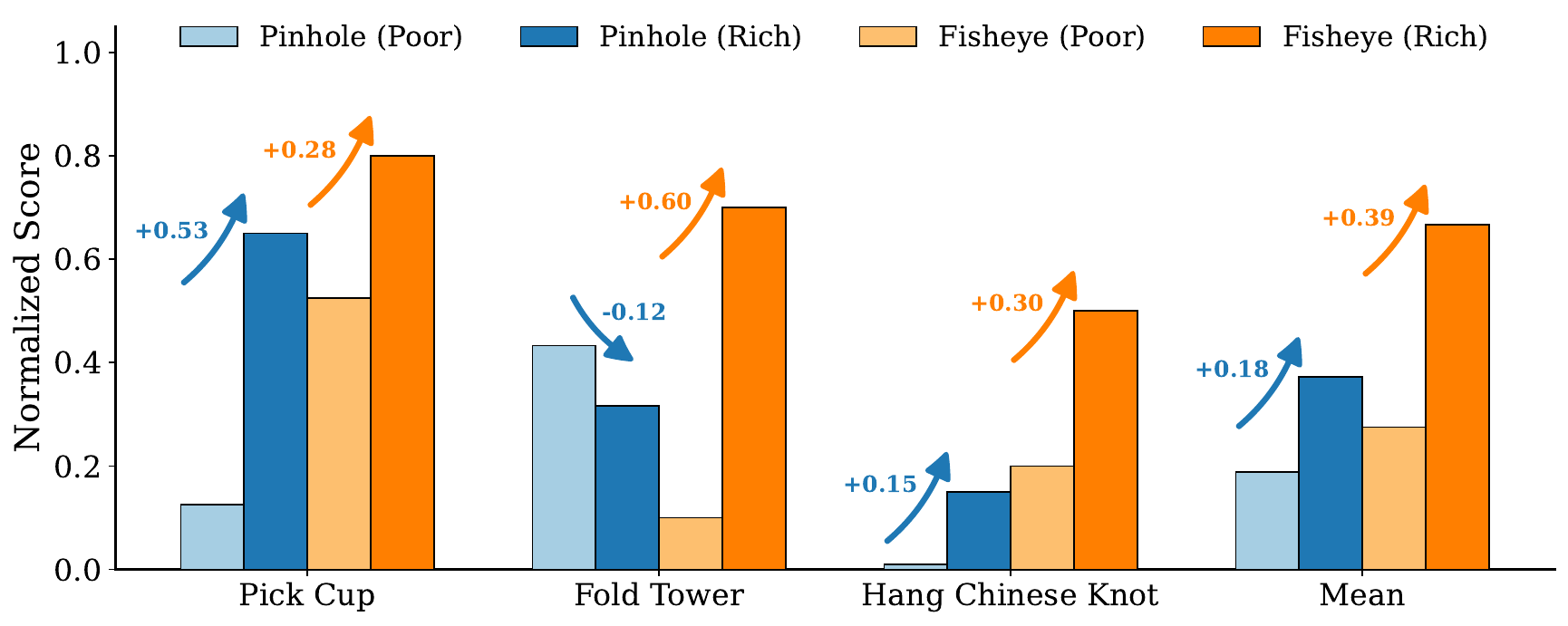}
   \caption{Real-world performance of fisheye / pinhole cameras with different scene complexity (\textit{feature-poor} v.s. \textit{feature-rich}) in three tasks (RQ1).}
   \label{fig:rq1_real}
   \vspace{-5mm}
\end{figure}

\section{Experimental Analysis}
\label{sec:experiments}

% --- RQ1 ---
\subsection{(RQ1) Spatial Localization}
\label{subsec:rq1_localization}
In this section, we investigate RQ1: \textit{Can the wider FoV of fisheye cameras help policy localization?}

Given our reliance on wrist-view-only vision, the policy must localize using background cues. We posit that the fisheye's advantage stems from its wider FoV capturing more static background features, which serve as stable visual anchors. This leads to our testable hypothesis:

\textbf{Hypothesis}: \textit{The fisheye's wider FoV enables superior policy localization by integrating a greater density of background features. Consequently, policy performance will exhibit a strong positive dependency on the visual richness of the training scene.}

To validate this, we conduct a two-part experiment. First, we measure task performance in ``feature-poor" (\eg uniform color) vs. ``feature-rich" (\eg complex background in \cref{fig:realworld_setup}(a)) backgrounds. The results in\cref{fig:rq1_real} and \cref{tab:rq1_simulation_bg_delta} confirm that rich backgrounds are critical. Crucially, \cref{fig:rq1_real} shows this performance gain is \textbf{more significant for the fisheye camera} in the real world (average gain $+0.39$ vs. $+0.18$ for pinhole) than in simulation. We attribute this to the stronger CLIP~\cite{clip} encoder and complex real-world textures, which the fisheye's FoV fully exploits.

Second, to prove this gain is caused by superior localization, we probe the encoder's implicit spatial awareness. We finetune the trained visual encoder with a lightweight MLP head to predict the robot's proprioceptive state (position/orientation) from images; lower error indicates better spatial awareness. The results in \cref{tab:rq1_proprioception_results} are conclusive. Encoders trained with fisheye cameras consistently yield lower proprioception error, and the \textbf{fisheye camera in a feature-rich environment} achieves the best performance by a large margin (\eg 1.73 cm translation error), confirming it learns the most accurate spatial representation.

Collectively, these findings confirm the fisheye's efficacy for localization is contingent on the environment. This provides a crucial guidance for data collection: \textbf{To maximize policy performance, data should be collected in visually complex and feature-rich environments to fully unlock the fisheye's spatial localization capabilities.}

\newcommand{\imp}[1]{\scriptsize(+#1)}
\newcommand{\dec}[1]{\scriptsize(-#1)}

\begin{table*}[!ht]
\caption{
    Simulation performance of fisheye / pinhole cameras with different scene complexity (\textit{feature-poor} v.s. \textit{feature-rich}) (RQ1).
    Performance in \textit{feature-rich }backgrounds is shown with the absolute difference (in parentheses) compared to the \textit{feature-poor }background baseline for the same camera. 
}
\label{tab:rq1_simulation_bg_delta}
\centering
\footnotesize
\setlength{\tabcolsep}{3pt}

\begin{tabular}{lc | c c c c c c | c}
\toprule
% --- 表头 ---
\multicolumn{2}{c |}{\textbf{Experimental Factors}} & \multicolumn{6}{c |}{\textbf{Simulation Task Success Rate}} & \multirow{2}{*}{\textbf{Average}} \\ 
\cmidrule(lr){1-2} \cmidrule(lr){3-8} 
\textsc{Camera} & \textsc{Feature} & \textsc{Square} & \textsc{Tool\_Hang} & \textsc{Coffee} & \textsc{Threading} & \textsc{Assemly} & \textsc{Mug\_clean} & \\ 
\midrule

Pinhole(Single) & {Poor} & 0.40 & 0.52 & 0.36  & 0.04 & 0.14 & 0.40 & 0.31 \\
Pinhole(Single) & {Rich} & 0.48 \imp{0.08} & 0.56 \imp{0.04} & 0.34 \dec{0.02} & 0.18 \imp{0.14} & 0.12 \dec{0.02} & 0.38 \dec{0.02} & 0.34 \imp{0.03} \\
\hdashline
\addlinespace[2pt]
Fisheye(Single) & {Poor} & 0.68 & 0.80 & \bfseries0.80  & 0.30 & 0.24 & 0.58 & 0.57 \\
Fisheye(Single) & {Rich} & \bfseries 0.74 \imp{0.06} & \bfseries 0.84 \imp{0.04} & 0.76 \dec{0.04} & \bfseries 0.56 \imp{0.26} & \bfseries 0.48 \imp{0.24} & \bfseries 0.60 \imp{0.02} & \bfseries 0.66 \imp{0.09} \\
\midrule
Pinhole(Double) & {Poor} & 0.50 & 0.44 & 0.26 & 0.22 & 0.44 & 0.40 & 0.38 \\
Pinhole(Double) & {Rich} & 0.70 \imp{0.20} &  0.34 \dec{0.10} & 0.36 \imp{0.10} & 0.38 \imp{0.16} & 0.34 \dec{0.10} & 0.56 \imp{0.16} & 0.45 \imp{0.07} \\
\hdashline
\addlinespace[2pt]
Fisheye(Double) & {Poor} & 0.86 & 0.84 & 0.74 & \bfseries 0.68 & \bfseries 0.56 & 0.66 & 0.72 \\
Fisheye(Double) & {Rich} & \bfseries 0.88 \imp{0.02} &  \bfseries 0.88 \imp{0.04} & \bfseries 0.86 \imp{0.12} &   0.66 \dec{0.02} &  0.44 \dec{0.12} & \bfseries 0.80 \imp{0.14} & \bfseries 0.75 \imp{0.03} \\
\bottomrule
\end{tabular}
\end{table*}

\begin{table}[h]
    \centering
    \footnotesize
    \setlength{\tabcolsep}{2pt}
    \caption{Quantitative probing of visual encoder spatial awareness using proprioception prediction as a proxy task (RQ1). We evaluate the quality of learned spatial representations by fine-tuning a lightweight MLP head on the pre-trained visual encoder to predict the robot's proprioceptive state in three real-world tasks.}
    \label{tab:rq1_proprioception_results}
    \begin{tabular}{lllcc}
        \toprule
        \textbf{Task} & \textbf{Camera} & \textbf{Feature} & \textbf{Trans. Err(cm)} $\downarrow$ & \textbf{Rot. Err($^\circ$)} $\downarrow$ \\
        \midrule

        % ---------------- Task 1: Pick Cup ----------------
        \multirow{4}{*}{\textbf{Pick Cup}} 
          & Pinhole & Poor & 12.309 & 15.345 \\
          & Pinhole & Rich & 5.367 & 7.612 \\
          & Fisheye & Poor & 3.369 & 3.677 \\
          & Fisheye & Rich & \textbf{2.362} & \textbf{3.394} \\
        \midrule

        % ---------------- Task 2: Fold Towel (Empty) ----------------
       \multirow{4}{*}{\textbf{Fold Towel}} 
          & Pinhole & Poor & 4.204 & 6.829 \\
          & Pinhole & Rich & 5.329 & 6.434 \\
          & Fisheye & Poor & 3.837 & 3.398 \\
          & Fisheye & Rich & \textbf{2.908} & \textbf{2.952} \\
        \midrule
       \multirow{4}{*}{\textbf{Hang Chinese Knot}} 
          & Pinhole & Poor & 14.168 & 12.464 \\
          & Pinhole & Rich & 7.683 & 9.377 \\
          & Fisheye & Poor & 8.804 & 7.256 \\
          & Fisheye & Rich & \textbf{5.143} & \textbf{4.887} \\
        \bottomrule

    \end{tabular}

\end{table}

% --- RQ2 ---
\subsection{(RQ2) Scene Generalization}
\label{subsec:rq2_background_gen}

This section investigates RQ2: \textit{How do fisheye cameras affect generalization to novel backgrounds?}

In robotic manipulation, the wrist-mounted camera's motion naturally induces background shifts, which can be viewed as a form of \textbf{implicit data augmentation} for scene generalization. We posit that the fisheye's wide FoV significantly amplifies this effect by capturing more extensive background changes with wider FoV and introducing stronger augmentation with the fisheye distortion during robot movement. This leads to our central hypothesis:

\textbf{Hypothesis}: \textit{Fisheye-trained policies can more effectively utilize scene diversity to improve generalization, exhibiting a steeper performance scaling curve as the number of unique training scenes increases.}

\begin{figure}[ht!]
  \centering
   \includegraphics[width=\linewidth]{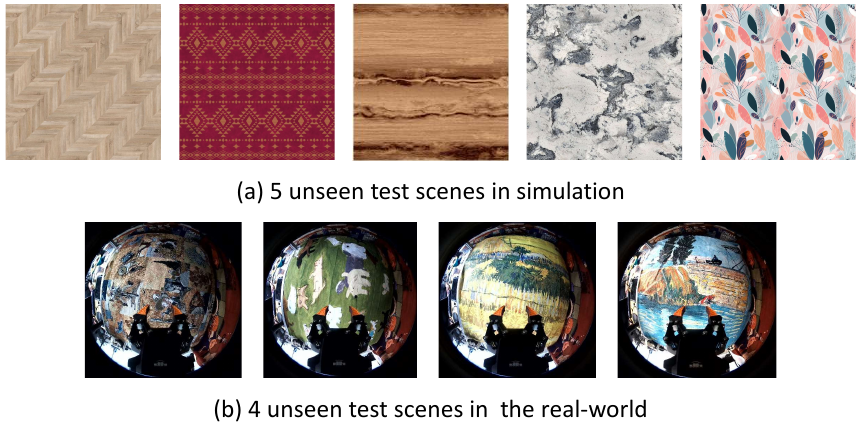}

   \caption{The unseen scenes for evaluation in (a) simulation and (b) real-world experiments (RQ2).}
   \label{fig:rq2_unseen_scenes}
   \vspace{-5mm}
\end{figure}

To verify this hypothesis, we design experiments to explicitly measure \textbf{the scaling effect of scene diversity}. We vary the number of unique training scenes, $N$, while holding the total data volume fixed, and evaluate the policy's zero-shot performance on distinct \textit{unseen} scenes in both simulation and the real world (see \cref{fig:rq2_unseen_scenes}). We systematically test $N=\{1, 8, 16, 32\}$ in simulation and $N=\{1, 2, 4, 6, 8\}$ in the real world.

\begin{figure}[ht!]
  \centering
   \includegraphics[width=\linewidth]{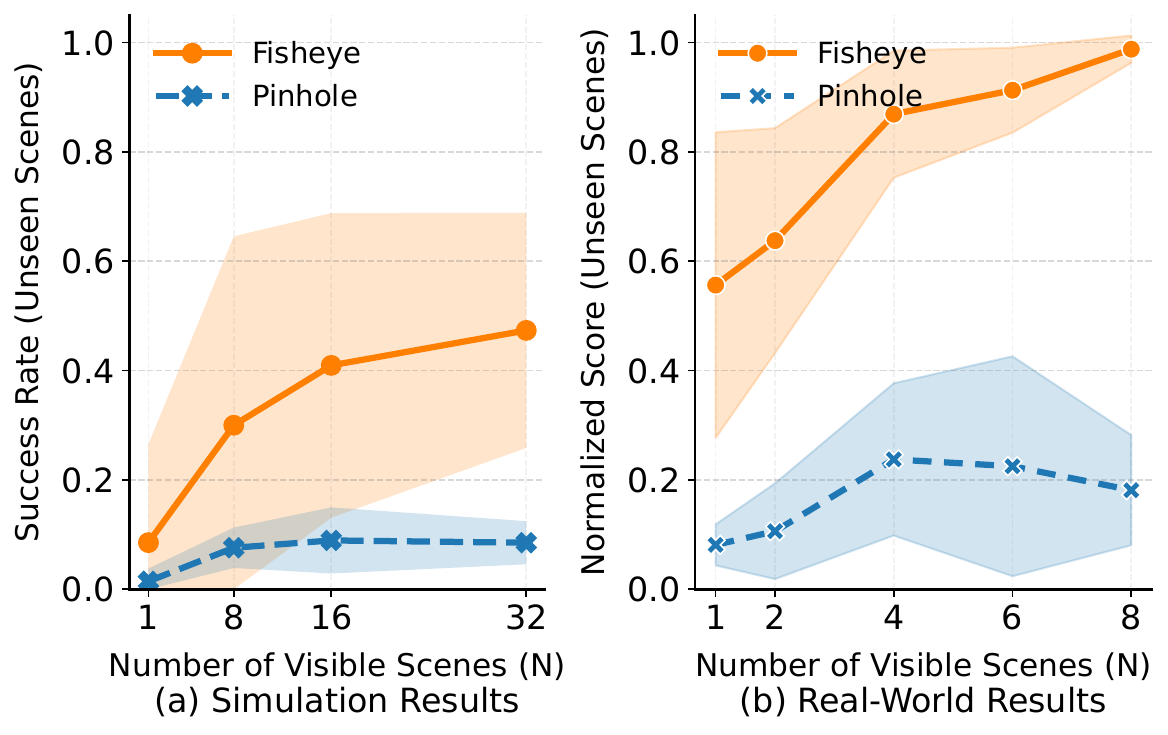}

   \caption{The policy performance improves with the number of training scenes in (a) simulation experiments on \textit{Coffee} task and (b) real-world experiments on \textit{Pick Cup} task (RQ2).}
   \label{fig:rq2_sim_real}
\end{figure}

The results, presented in \cref{fig:rq2_sim_real}, strongly support our hypothesis. We observe that the fisheye camera exhibits significantly greater scaling potential compared to the conventional camera. Notably, in the real-world setup, the fisheye policy's zero-shot success rate on unseen environments rapidly exceeds $95\%$ when trained with just eight diverse scenes. In contrast, the scaling curve in simulation is less steep. We attribute this discrepancy to two primary differences: 1) the variation in visual encoders (a non-pre-trained ResNet-18~\cite{resnet} in simulation vs. the pre-trained CLIP~\cite{clip} in the real world), and 2) the comparatively lower visual complexity of simulated background imagery (see \cref{fig:realworld_setup} and the supplementary file).

Collectively, these findings confirm that the wider FoV of the fisheye camera acts as a potent implicit data augmentation, enabling the policy to better leverage scene diversity for robust cross-scene generalization. This yields a crucial guidance for large-scale data collection: \textbf{maximizing scene diversity is essential to unlock the full generalization capabilities of fisheye cameras.}

% --- RQ3 ---
\subsection{(RQ3) Hardware Generalization}
\label{subsec:rq3_hardware_gen}

This section investigates RQ3: \textit{Can policies maintain performance when deployed on new fisheye lenses?}

This question is of significant practical importance. First, as large-scale, fisheye-based datasets (\eg for pre-training VLAs) become more common~\cite{huDataScalingLaws2024,rdt2,generalist2025gen0}, it is crucial that policies can be fine-tuned and deployed on custom robotic setups, which will inevitably use lenses with different intrinsic parameters. Second, as hardware is upgraded over time, policies must remain backward-compatible with legacy data. However, the pronounced and varied distortion profiles of fisheye lenses make this cross-camera transfer a non-trivial challenge.

For wrist-mounted cameras, we observe that policies heavily rely on spatial relationships (\eg object distance) from the \textbf{absolute scale} of objects and the gripper in the image. When a new lens is introduced, these absolute scales change, causing the policy to misinterpret the scene (\eg perceiving an object as closer or farther than it is) and leading to catastrophic failures (see \cref{fig:rq3_failure_case}). This leads to our key hypothesis:

\begin{figure}[ht!]
  \centering
   \includegraphics[width=0.7\linewidth]{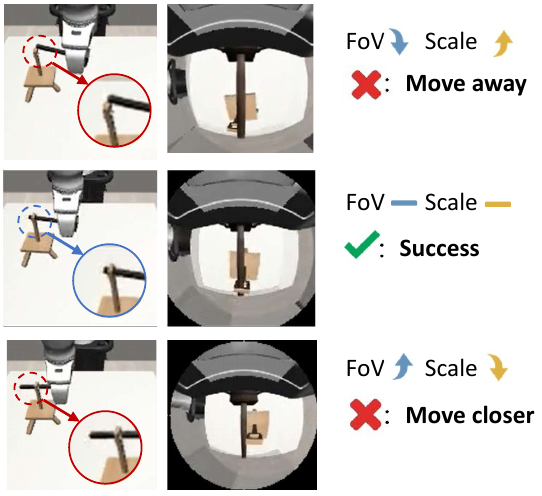}

   \caption{The failure cases of cross-hardware generalization (RQ3).
   The policy tends to overfit the absolute scale.
   }
   \label{fig:rq3_failure_case}
\end{figure}

\textbf{Hypothesis}: \textit{The primary challenge of cross-lens transfer is policy overfitting to absolute object scales. This can be mitigated by using augmentation to force the policy to focus on \textbf{relative scale relationships} (\eg object scale relative to the gripper) instead.}

As introduced in \cref{subsec:enabling_components}, we propose \textbf{Random Scale Augmentation (RSA)}, a simple yet effective strategy to address this problem (see \cref{fig:scale_augmentation}). RSA forces the network to observe the target object and gripper fingers at diverse relative scales, breaking its reliance on absolute pixel size.

To verify this hypothesis, we conduct extensive experiments in simulation, as evaluating numerous hardware configurations in the real world is costly (see supplementary for real-world verification). We train policies on a single camera configuration (``Seen Param") and evaluate their zero-shot transfer performance on five \textit{unseen} configurations with varying distortion and FoV parameters (see supplementary for more details).

The results in \cref{fig:rq3_sim} are conclusive. The baseline policy, trained with normal augmentations, exhibits a severe performance drop when deployed on unseen lenses (\eg ``Param 3", ``Param 4"). In contrast, the policy trained with our RSA maintains higher success rates across all configurations, demonstrating robust generalization. This strongly supports our hypothesis that learning relative scale is the key to cross-camera robustness.

This analysis yields our final guidance for fisheye data collection: \textbf{Employ strong scale-based augmentation like RSA to ensure policies are robust to hardware variations and can leverage data from diverse lens sources.}
\begin{figure}[ht!]
  \centering
   \includegraphics[width=\linewidth]{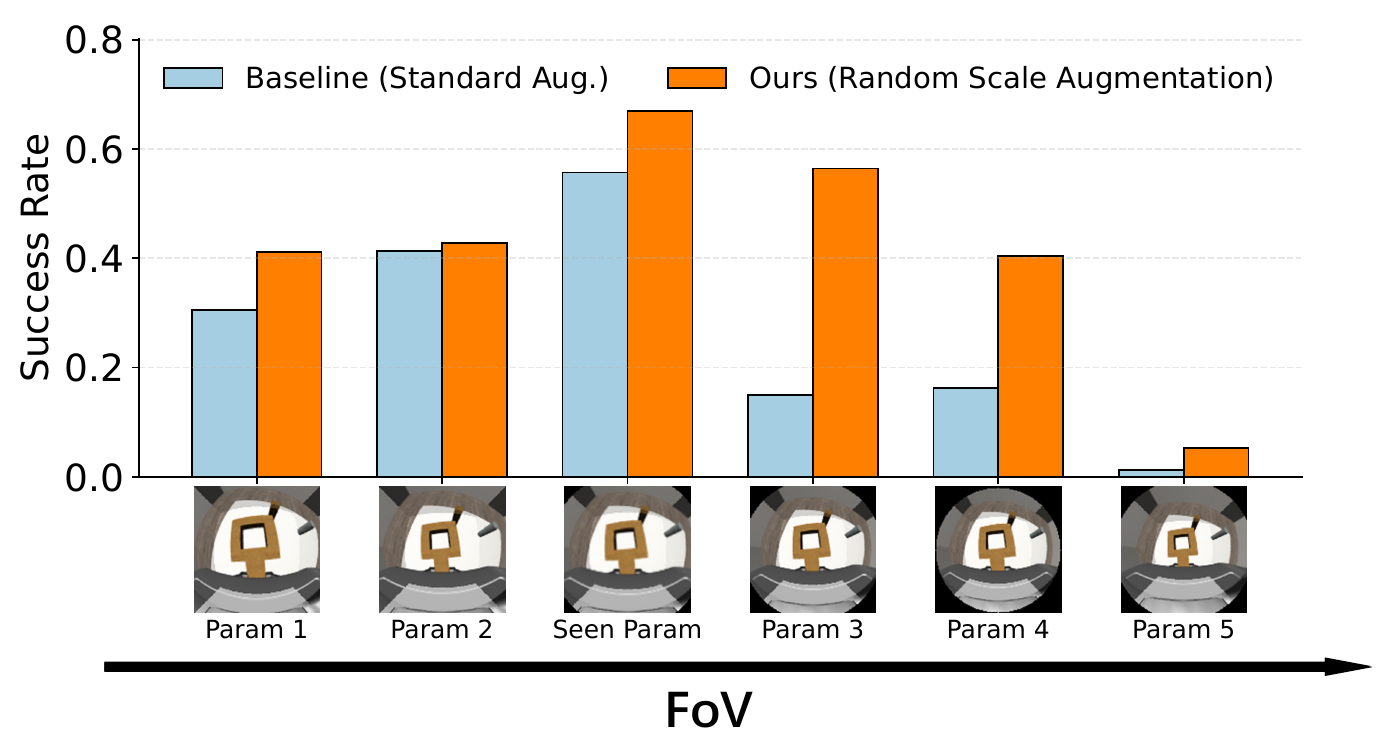}

   \caption{The policy performance under different \textit{unseen} camera parameters , averaged across six tasks in simulation  (RQ3).}
   \label{fig:rq3_sim}
   \vspace{-5mm}
\end{figure}
\section{Conclusion}
We presented the first systematic analysis of wrist-mounted fisheye cameras for imitation learning, investigating spatial localization, scene, and hardware generalization. Our key findings are: (1) The wide FoV's localization benefit is critically contingent on environmental feature richness. (2) Fisheye cameras unlock superior scene generalization by better leveraging data diversity. (3) Cross-camera transfer failure, caused by scale overfitting, is effectively solved by our proposed Random Scale Augmentation. These findings provide concrete, actionable guidelines for the robotics community, offering an empirical foundation for large-scale data collection and the training of robust, generalist robot policies with fisheye cameras.

{
    \small
    \bibliographystyle{ieeenat_fullname}
    \bibliography{main}
}

\clearpage
\appendix

\renewcommand{\thesection}{\Alph{section}}
\renewcommand{\thefigure}{A\arabic{figure}}
\renewcommand{\thetable}{A\arabic{table}}
\setcounter{section}{0}
\setcounter{figure}{0}
\setcounter{table}{0}

\clearpage 
\appendix

\twocolumn[
  \begin{center}
    \textbf{\Large Supplementary Material for \\ \vspace{5pt} Rethinking Camera Choice: An Empirical Study on Fisheye Camera Properties in Robotic Manipulation}
    \vspace{24pt}
  \end{center}
]

\setcounter{table}{0}
\renewcommand{\thetable}{S\arabic{table}}
\setcounter{figure}{0}
\renewcommand{\thefigure}{S\arabic{figure}}
\setcounter{equation}{0}
\renewcommand{\theequation}{S\arabic{equation}}
\section{Overview}
\label{sec:supp_overview}
This supplementary material provides comprehensive implementation details, extended analyses, and additional real-world verifications to substantiate the findings presented in the main paper. The material is organized as follows:

\begin{itemize}
    \item \textbf{Section~\ref{sec:supp_env_visualizations}} (\textbf{Experiment Setup Details}) focuses on the experiment setup details, covering the visualization of each simulation task, the double camera setup, and the comparison of experimental scenes.
    \item \textbf{Section~\ref{sec:supp_imp_details}} (\textbf{Implementation Details}) provides the specific hyperparameters for training and the definitions of normalized score for real-world experiments.
    
    \item \textbf{Section~\ref{sec:supp_rq1}} (\textbf{RQ1: Spatial Localization}) details the probing model and presents additional ablation studies on proprioception.
    
    \item \textbf{Section~\ref{sec:supp_rq2}} (\textbf{RQ2: Scene Generalization}) details the scene datasets used in our experiments and provides granular curves for both simulation and real-world tasks.
    
    \item \textbf{Section~\ref{sec:supp_rq3}} (\textbf{RQ3: Hardware Generalization}) specifies the camera parameters in simulation and presents the quantitative results of cross-camera experiments in simulation and the real-world. 
\end{itemize}

\section{Experiment Setup Details}
\label{sec:supp_env_visualizations}

\subsection{Simulation Tasks}
We show all the simulation tasks in~\cref{fig:sim_task}. They span a wide variety of behaviors including pick-and-place, precise insertion(e.g., Threading and Square), and include long-horizon tasks requiring chaining several behaviors together(e.g., Tool Hang and Mug Cleanup).
The detailed configuration, including trajectory counts and data sources for each task, is provided in~\cref{tab:sim_task_info}.

\begin{table}[h]
\centering
\caption{\textbf{Simulation Tasks Overview.} PH: Proficient-Human datasets; D0: Default reset distribution; D1: Broadened reset distribution.}
\label{tab:sim_task_info}
\vspace{2mm}
\resizebox{1.0\linewidth}{!}{
\begin{tabular}{lccc}
\toprule
\textbf{Task} & \textbf{Trajectory Counts} & \textbf{Data Source} & \textbf{Data Type} \\
\midrule
Square & 200 & RoboMimic & PH \\
Tool Hang & 200 & RoboMimic & PH \\
Coffee & 500 & MimicGen & D1 \\
Threading & 500 & MimicGen & D0 \\
Assembly & 500 & MimicGen  & D0 \\
Mug Cleanup & 500 & MimicGen & D0 \\
\bottomrule
\end{tabular}
}
\end{table}

\begin{figure}[htbp!]
  \centering
   \includegraphics[width=0.8\linewidth]{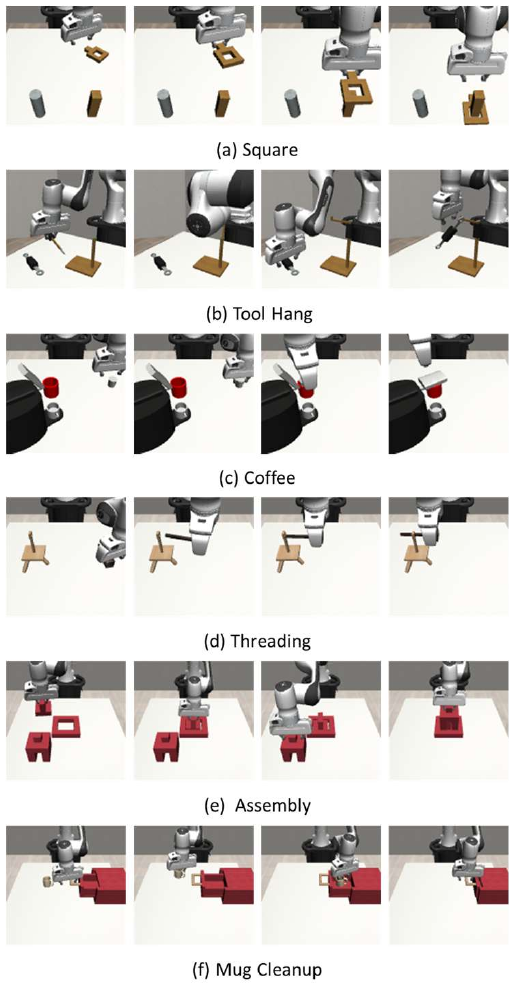}
   \caption{\textbf{Examples of simulation tasks.} Each row illustrates sequential snapshots of a distinct task: (a) Square, grasp the yellow square block and insert it into the yellow target slot. (b) Tool Hang, insert the needle-shaped hook and hang the tool. (c) Coffee, place the coffee pod into the machine and close the lid. (d) Threading, grasp the needle and insert it into the pinhole. (e) Assembly, insert two irregular blocks in sequence. (f) Mug Cleanup, open the drawer, place the mug inside, and close it.}
   \label{fig:sim_task}
\end{figure}

\subsection{Double Camera Setup in Simulation}
\begin{figure}[htbp!]
  \centering
   \includegraphics[width=0.8\linewidth]{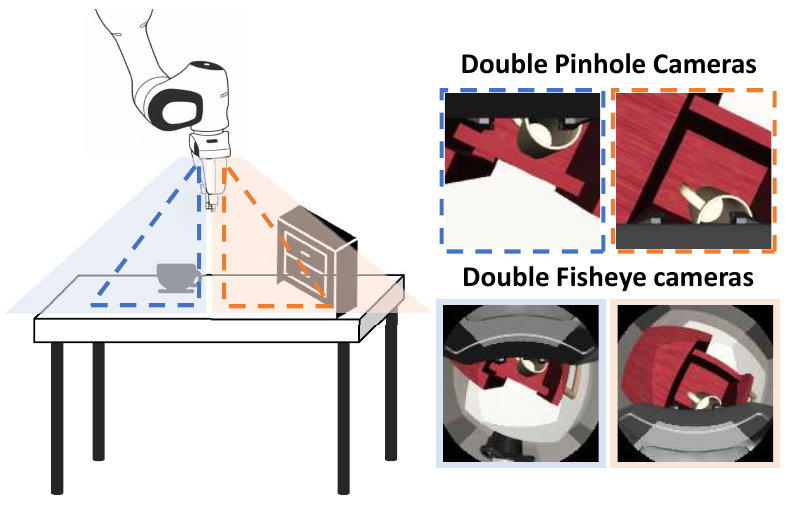}
   \caption{\textbf{Visualization of different cameras}. The orange area represents the field of view of the pinhole wrist camera, and the blue area represents the field of view of the newly added symmetrically arranged camera. The upper part shows the images from the pinhole wrist camera, and the lower part shows the images from the double fisheye cameras, with blue and orange borders distinguishing the imaging effects corresponding to their respective fields of view.}
   \label{fig:double_camera}
\end{figure}

To expand the field of view for more comprehensive scene perception, we added a new camera opposite to the original wrist camera in the simulation environment. 
The original wrist camera has a position parameter of pos=``$0.05$ 0 0'', and the newly added camera is placed on its opposite side with a position parameter of pos=``$-0.05$  0 0''. 
They are symmetrically distributed, thus achieving an effective expansion of the field of view.
The visualization of different cameras are shown in~\cref{fig:double_camera}

\section{Implementation Details}
\label{sec:supp_imp_details}

\subsection{Training Hyperparameters}
\label{subsec:hyperparams}

Our policy implementation is built upon the Diffusion Policy framework~\cite{chi2024diffusionpolicy}. Following the protocol established in~\cite{huDataScalingLaws2024}, we ensure rigorous control over hyperparameters to allow for fair comparisons between fisheye and pinhole cameras.

The specific hyperparameters for simulation and real-world experiments are detailed in \cref{tab:hyperparams}. 

\begin{table}[h]
    \centering
    \caption{\textbf{Detailed hyperparameters.} We report the specific settings used for Simulation (Sim) and Real-World (Real) experiments. The hyperparameter values are aligned with~\cite{chi2024diffusionpolicy} and~\cite{huDataScalingLaws2024}.}
    \label{tab:hyperparams}
    \vspace{2mm}
    \resizebox{1.0\linewidth}{!}{ 
    \begin{tabular}{lcc}
        \toprule
        \textbf{Config} & \textbf{Simulation} & \textbf{Real-World} \\
        \midrule
        \multicolumn{3}{l}{\textit{\textbf{Model Architecture}}} \\
        Visual Backbone & ResNet-18 (No Pretrain) & CLIP ViT-B/16\\
        Pooling Method & Spatial Softmax & Spatial Softmax \\
        Denoising Network & Conditional U-Net1D & Conditional U-Net1D \\
        Action Space & Relative Action & Relative Action \\
        Action Horizon ($T_{act}$) & 8 & 8 \\
        Observation Horizon ($T_{obs}$) & 2 & 2 \\
        Prediction Horizon ($T_{pred}$) & 16 & 16 \\
        \midrule
        \multicolumn{3}{l}{\textit{\textbf{Input Data}}} \\
        Image Resolution & $ 128 \times 128$ & $224 \times 224$ \\
        Image Preprocessing & Random Crop & Random Crop \\
        Proprioceptive Input & \textbf{None} (State-free) & \textbf{None} (State-free) \\
        \midrule
        \multicolumn{3}{l}{\textit{\textbf{Optimization}}} \\
        Optimizer & AdamW & AdamW \\
        Weight Decay & $1 \times 10^{-6}$ & $1 \times 10^{-6}$ \\
        LR Schedule & Cosine Decay & Cosine Decay \\
        Learning Rate (UNet) & $1 \times 10^{-4}$ & $1 \times 10^{-4}$ \\
        Learning Rate (Encoder) & $1 \times 10^{-4}$ & $1 \times 10^{-5}$ \\
        Batch Size & 16 & 64 \\
        Training Epochs & 2000 & 500 \\
        EMA Decay & 0.75 & 0.9999 \\
        \bottomrule
    \end{tabular}
    }
\end{table}

\subsection{Real-World Task Score Metric Definitions}
\label{subsec:real_task_def}
As described in the main paper, we employ a normalized, multi-stage scoring metric for real-world evaluation. This metric provides a more granular assessment of policy capability than binary success rates. The detailed breakdown for each task is defined below:

\begin{itemize}
    \item \textbf{Pick Cup}: The goal is to pick up a cup and place it upright onto a coaster.
    \begin{itemize}
        \item \textit{Stage 1 (0.00 pts):} Failed to grasp the cup, or grasped the cup but failed to place it onto the coaster (e.g., dropped it or missed the target).
        \item \textit{Stage 2 (0.50 pts):} Successfully grasped and placed the cup onto the coaster, but the cup toppled over (not upright).
        \item \textit{Stage 3 (1.00 pts):} Successfully grasped and placed the cup onto the coaster, maintaining an upright orientation.
    \end{itemize}

   \item \textbf{Fold Towel}: The goal is to perform two consecutive folds on a towel. The robot must grasp a corner, fold it to the diagonal line, and then grasp the other corner to complete the second fold.
    \begin{itemize}
        \item \textit{Stage 1 (0.25 pts):} Successfully grasped the first corner and lifted it off the table surface.
        \item \textit{Stage 2 (0.50 pts):} Successfully released the gripper and completed the first fold.
        \item \textit{Stage 3 (0.75 pts):} Successfully localized and grasped the second corner (after the first fold) and lifted it off the table.
        \item \textit{Stage 4 (1.00 pts):} Successfully released the gripper and completed the second fold.
    \end{itemize}

   \item \textbf{Hang Chinese Knot}: The goal is to hang a Chinese knot onto a designated hook on a stand. Since the initial grasping phase is successfully completed by most baselines, we do not include it as a scoring criterion. We focus solely on the precise placement required to secure the knot.
    \begin{itemize}
        \item \textit{Stage 1 (0.00 pts):} Failed to hang the knot onto the hook (e.g., dropped midway or missed the hook).
        \item \textit{Stage 2 (1.00 pts):} Successfully moved the knot to the target location and secured the knot onto the hook.
    \end{itemize}
\end{itemize}

For each evaluation setup, we conduct 20 trials and report the \textbf{cumulative score} (sum of scores across all trials).

\section{Additional Experiments for RQ1 (Spatial Localization)}
\label{sec:supp_rq1}

In this section, we first show the scenes in RQ1 simulation experiments (\cref{experiment-scene-sim}). Then we elaborate on the model details for probing the spatial awareness of the visual encoders (\cref{subsec:proprio_probing}). Additionally, to further validate our findings on spatial localization, we evaluate the policy's performance with proprioceptive input (\cref{ablation:simulation-with-pro} and \cref{ablation:policy-performance}) and third-view input (\cref{ablation:third-view}).

\subsection{Experimental Scenes in Simulation}
\label{experiment-scene-sim}
\begin{figure}[htbp!]
  \centering
   \includegraphics[width=\linewidth]{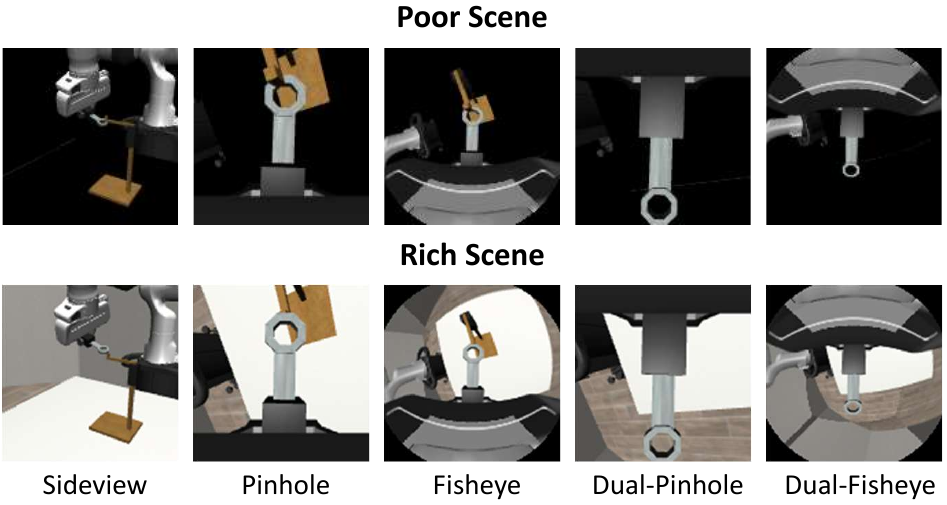}

   \caption{\textbf{Scene comparison in RQ1.} Visualization of sideview, pinhole, fisheye, dual-pinhole, and dual-fisheye cameras in poor (single dark) vs. rich (diverse elements) scenes.}
   \label{fig:sim_rq1_scene}
\end{figure}

The~\cref{fig:sim_rq1_scene} presents a scene example from the Tool Hang task in the RQ1 experiment, comparing visualization effects across different cameras including third-view camera, pinhole camera, and fisheye camera under two distinct environmental settings: a poor scene (single dark scene) and a rich scene (scene with diverse elements).

\subsection{Implementation Details: Proprioception Prediction Task}
\label{subsec:proprio_probing}
\textbf{Methodology.} 
In the main paper, we demonstrated that fisheye-based policies achieve lower proprioception prediction errors in real-world tasks. Here, we provide the detailed experimental setup for this probing task. To explicitly quantify the spatial information captured by the visual representations, we fine-tune the visual encoder on a proprioception prediction task. 
Specifically, we extract the visual encoder (CLIP ViT~\cite{clip}) from the learned manipulation policy and attach a lightweight Multi-Layer Perceptron (MLP) head. The MLP consists of two fully connected layers with 256 hidden units and ReLU activations. We fine-tune the entire network (encoder + MLP head) to regress the robot's end-effector pose (3D position and quaternion orientation) using Mean Squared Error (MSE) loss. To rigorously evaluate the generalization of the learned spatial representations, we do not simply split the training data. Instead, we collected a dedicated \textbf{test set comprising 30 additional trajectories} for each task setup. These trajectories were collected under the exact same environmental settings as the training data but were kept strictly held-out during the training phase.

\subsection{Ablation Study: Simulation Policy Performance with Proprioception}
\label{ablation:simulation-with-pro}
\textbf{Motivation.} As detailed in Section 3.3, we deliberately excluded proprioceptive state (e.g., end-effector pose and joint positions) from the policy input to rigorously isolate and evaluate the visual spatial localization capabilities of different camera models (RQ1). However, in practical robotic applications, proprioception is often available. In this section, we first conduct an ablation study in simulation to investigate how the addition of proprioceptive state affects policies trained with Pinhole and Fisheye cameras. All experiments in this section are conducted under the ``Feature-Rich'' background setting to ensure fair comparison.

\textbf{Simulation Experimental Setup.}
All experiments in this section are conducted under the ``Feature-Rich'' background setting to ensure a fair comparison. We evaluate performance across six simulation tasks. For brevity in \cref{tab:sim_ablation_proprio}, we use the following abbreviations: \textbf{Tool} (Tool Hang), \textbf{Sqr} (Square), \textbf{Cof} (Coffee), \textbf{Thrd} (Threading), \textbf{Asm} (Assembly) and \textbf{Mug} (Mug Cleanup).
We compare two wrist-camera configurations:
\begin{itemize}
    \item \textbf{Single:} The robot is equipped with a single wrist-mounted camera.
    \item \textbf{Double:} The robot is equipped with two wrist-mounted cameras (providing different views) to reduce occlusion.
\end{itemize}

\begin{table*}[t]
\centering
\footnotesize
\caption{\textbf{Ablation study on proprioception input in simulation.} We compare the success rates between (\textit{w/ State}) and without (\textit{w/o State}) proprioceptive input. Values in $(\cdot)$ indicate the performance drop (or gain) relative to the baseline \textit{w/ State} performance. Pinhole cameras show a sharp relative decline (e.g., -41\% performance drop on average), whereas Fisheye cameras maintain robust performance (e.g., only -6\% drop on average).}

\setlength{\tabcolsep}{3.5pt}
\begin{tabular}{lc|cccccc|c}
\toprule
\multicolumn{2}{c}{\textbf{Experimental Factors}} & \multicolumn{6}{c}{\textbf{Simulation Task Success Rate}} & \multirow{2}{*}{\textbf{Average}} \\
\cmidrule(lr){1-2} \cmidrule(lr){3-8}
\textsc{Camera} & \textsc{State} & \textsc{Square} & \textsc{Tool\_Hang} & \textsc{Coffee} & \textsc{Threading} & \textsc{Assembly} & \textsc{Mug\_Clean} & \\
\midrule

\multirow{2}{*}{Pinhole(Single)} 
 & w/ State & 0.82 & 0.68 & 0.38 & 0.72 & 0.46 & 0.66 & 0.62 \\
 & w/o State & 0.48 {(-41\%)} & 0.56 {(-18\%)} & 0.34 {(-11\%)} & 0.18 {(-75\%)} & 0.12 {(-74\%)} & 0.38 {(-42\%)} & 0.34 {(-45\%)} \\
\cmidrule(lr){1-8} \cmidrule(lr){9-9}

\multirow{2}{*}{Fisheye(Single)} 
 & w/ State & 0.86 & 0.88 & 0.88 & 0.72 & 0.58 & 0.60 & 0.75 \\
 & w/o State & \textbf{0.74} {(-14\%)} & \textbf{0.84} {(-5\%)} & \textbf{0.76} {(-14\%)} & \textbf{0.56} {(-22\%)} & \textbf{0.48} {(-17\%)} & \textbf{0.60} {(0\%)} & \textbf{0.66} {(-12\%)} \\
\midrule
\multirow{2}{*}{Pinhole(Double)} 
 & w/ State & 0.92 & 0.54 & 0.38 & 0.76 & 0.58 & 0.66 & 0.64 \\
 & w/o State & 0.70 {(-24\%)} & 0.34 {(-37\%)} & 0.36 {(-5\%)} & 0.38 {(-50\%)} & 0.34 {(-41\%)} & 0.56 {(-15\%)} & 0.45 {(-30\%)} \\
\cmidrule(lr){1-8} \cmidrule(lr){9-9}

\multirow{2}{*}{Fisheye(Double)} 
 & w/ State & 0.94 & 0.88 & 0.88 & 0.78 & 0.56 & 0.80 & 0.81 \\
 & w/o State & \textbf{0.88} {(-6\%)} & \textbf{0.88} {(0\%)} & \textbf{0.86} {(-2\%)} & \textbf{0.66} {(-15\%)} & \textbf{0.44} {(-21\%)} & \textbf{0.80} {(0\%)} & \textbf{0.75} {(-7\%)} \\
\bottomrule
\end{tabular}
\label{tab:sim_ablation_proprio}
\end{table*}

\textbf{Results and Analysis.}
The quantitative results are summarized in \cref{tab:sim_ablation_proprio}. The \textbf{Average} column clearly illustrates the divergent reliance on proprioception between the two camera settings:

\begin{enumerate}
    \item \textbf{Pinhole Sensitivity:} Pinhole-based policies suffer a catastrophic performance drop when proprioception is removed. For the \textit{Single Pinhole} configuration, the mean success rate plummets from 0.62 to 0.34 (a drop of \textbf{45\%}). For the \textit{double Pinhole} configuration, the mean success rate plummets from 0.64 to 0.45 (a drop of \textbf{30\%}). This confirms that without the explicit guidance of robot state, the narrow FoV struggles to maintain consistent localization.
    
    \item \textbf{Fisheye Robustness:} In sharp contrast, Fisheye-based policies exhibit remarkable robustness. The \textit{Single Fisheye} configuration maintains a high mean success rate (dropping only slightly from 0.75 to \textbf{0.66}), and the \textit{Double Fisheye} setup sees a negligible decline (0.81 to \textbf{0.75}). This empirically proves that the fisheye's wide contextual view implicitly encodes the robot's spatial relationship with the environment effectively, rendering explicit state input largely redundant.
\end{enumerate}

\subsection{Ablation Study: Policy Performance with Proprioception}
\label{ablation:policy-performance}
\textbf{Real-World Experimental Setup.} To validate our simulation findings in the real-world, we conducted the same ablation study using the Real-World setup described in the main paper. We utilized the ``Feature-Rich'' (Changeable Background with rich textures) setting to maximize the potential for visual feature extraction. We compare the Normalized Score of the policy with and without proprioception across three real-world tasks: \textbf{Pick Cup}, \textbf{Fold Towel}, and \textbf{Hang Chinese Knot}.

\begin{table}[h]
\centering
\footnotesize
\caption{\textbf{Real-World ablation on proprioception.} We report the Normalized Score in the feature-rich setting. Values in $(\cdot)$ indicate the performance drop relative to the baseline \textit{w/ State} performance. Notably, the Fisheye policy without proprioception (0.67) outperforms the Pinhole policy even with proprioception (0.52).}
\label{tab:real_ablation_proprio}
\setlength{\tabcolsep}{0.5pt}
\begin{tabular}{lc|ccc|c}
\toprule
\multicolumn{2}{c}{\textbf{Experimental Factors}} & \multicolumn{3}{c}{\textbf{Real-World Normalized Score}} & \multirow{2}{*}{\textbf{Average}} \\
\cmidrule(lr){1-2} \cmidrule(lr){3-5}
\textsc{Camera} & \textsc{State} & \textsc{Pick Cup} & \textsc{Fold Towel} & \textsc{Hang Knot} & \\
\midrule

\multirow{2}{*}{Pinhole} 
 & w/ State & 0.75 & 0.37 & 0.45 & 0.52 \\
 & w/o State & 0.65 {(-13\%)} & 0.32 {(-14\%)} & 0.15 {(-67\%)} & 0.37 {(-29\%)} \\
\midrule

\multirow{2}{*}{Fisheye} 
 & w/ State & 0.98 & 0.92 & 0.70 & 0.87 \\
 & w/o State & \textbf{0.80} {(-18\%)} & \textbf{0.70} {(-24\%)} & \textbf{0.50} {(-29\%)} & \textbf{0.67} {(-23\%)} \\
\bottomrule
\end{tabular}
\end{table}

\textbf{Results.} The real-world results, presented in \cref{tab:real_ablation_proprio}, reveal an even more pronounced advantage for fisheye cameras compared to simulation:
\begin{enumerate}
    \item \textbf{Superior Spatial Localization:} The most critical metric is the performance \textit{without} proprioception (w/o State), which represents the camera's pure visual localization capability. Fisheye cameras significantly outperform Pinhole cameras in this regime. For instance, in the challenging deformable object task (\textit{Fold Towel}), the Fisheye policy achieves a score of \textbf{0.70} without state, whereas the Pinhole policy struggles at \textbf{0.32}. On average, the Fisheye camera achieves a mean score of \textbf{0.67} using only vision, nearly doubling the Pinhole camera's mean of \textbf{0.37}.
    
    \item \textbf{Reduced State Dependency:} While adding proprioception improves performance for both cameras (likely due to the inherent noise and dynamics of the real world), Pinhole cameras are far more dependent on it. In the \textit{Hang Chinese Knot} task, the Pinhole policy relies on state to jump from a failing score of 0.15 to 0.45. In contrast, the Fisheye policy already starts at a strong baseline of 0.50 purely from vision.
\end{enumerate}

\vspace{0.5em}
\noindent \textbf{Summary of Ablation.} These real-world experiments, consistent with our simulation findings, corroborate our hypothesis: the wide FoV of the fisheye camera captures sufficient global context to enable high-precision manipulation even in the absence of robot state information. Collectively, these results reinforce our conclusion in RQ1 that fisheye cameras inherently provide superior spatial localization capabilities, significantly reducing the dependency on precise robot state input.

\subsection{Ablation Study: Impact of Third-view Camera Integration}
\label{ablation:third-view}
While the primary study isolates the effects of wrist-mounted cameras by excluding additional sensors, real-world robotic deployment often incorporates multi-modal setups, such as combining wrist cameras with proprioception or third-person views. To investigate how fisheye cameras behave in these more complex sensing paradigms, we evaluated a "Wrist + Third-person" configuration in simulation. Although recent state-of-the-art frameworks like UMI\cite{chi2024universal} and GEN-0\cite{generalist2025gen0} primarily rely on wrist cameras, our exploration provides informative insights for broader deployment scenarios.
\begin{table}[h]
\centering
\scriptsize
\setlength{\tabcolsep}{2pt}
\caption{Third-view Camera Ablation  in Simulation.}
\label{tab:third_view}
\begin{tabular}{lccccccc}
\toprule
\textbf{Config (Double Cam + 3rd)} & \textbf{Sqr} & \textbf{Tool} & \textbf{Cof} & \textbf{Thrd} & \textbf{Asm} & \textbf{Mug} & \textbf{Mean} \\ \midrule
Pinhole baseline & 0.94 & 0.78 & 0.78 & 0.80 & 0.56 & 0.66 & 0.75 \\
Fisheye (\textbf{Ours}) & \textbf{0.96} & \textbf{0.84} & \textbf{0.82} & \textbf{0.82} & \textbf{0.66} & \textbf{0.72} & \textbf{0.80} \\
\textit{Improvement} & \textit{+0.02} & \textit{+0.06} & \textit{+0.04} & \textit{+0.02} & \textit{+0.10} & \textit{+0.06} & \textit{\textbf{+0.05}} \\
\bottomrule
\end{tabular}
\end{table}\\
As summarized in \cref{tab:third_view}, the fisheye camera consistently maintains a 5\% mean performance gain over the pinhole baseline even when a third-person view is available. Notably, in the high-precision "Asm" (Assembly) task, the fisheye configuration achieves a 10\% improvement, demonstrating that the wide-FoV benefits of fisheye cameras are not redundant when global views are present. Instead, fisheye lenses provide essential local context—such as precise gripper-object relative poses—that fixed global cameras may struggle to capture due to occlusions or limited resolution. This consistent gain proves that the advantages of fisheye cameras identified in our study are robust and carry over to more comprehensive sensor suites, further justifying their adoption in future generalist robot policies.
\section{Additional Experiments for RQ2 (Scene Generalization)}
\label{sec:supp_rq2}

\subsection{Visualization of Environmental Diversity}

\textbf{Motivation.} 
In the main paper, we established that the wide FoV of fisheye cameras significantly enhances spatial localization, particularly in feature-rich environments (RQ1), and that this capability scales with scene diversity (RQ2). To validate that our experimental setup provides a rigorous assessment of generalization rather than simple memorization, we provide both a qualitative visualization and a quantitative distribution analysis of the background datasets.

\begin{figure*}[t]
    \centering
    \begin{minipage}{\textwidth}
        \centering
        \includegraphics[width=0.77\linewidth]{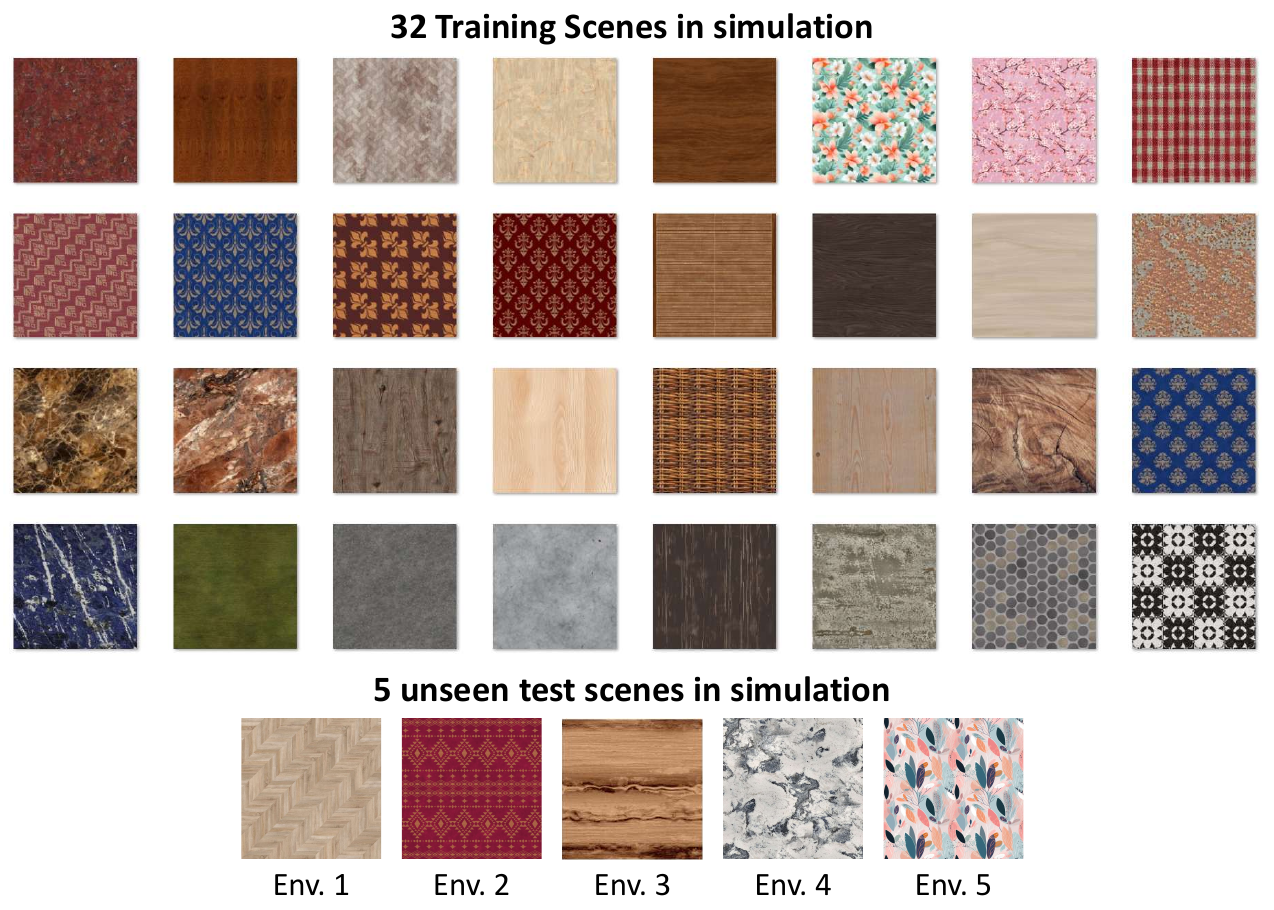} 
        \vspace{-5pt}
        \centerline{\small (a) Simulation Texture Library: High-frequency textures providing randomized visual noise in simulation.}
    \end{minipage}
    
    \vspace{6pt}

    \begin{minipage}{\textwidth}
        \centering
        \includegraphics[width=0.63\linewidth]{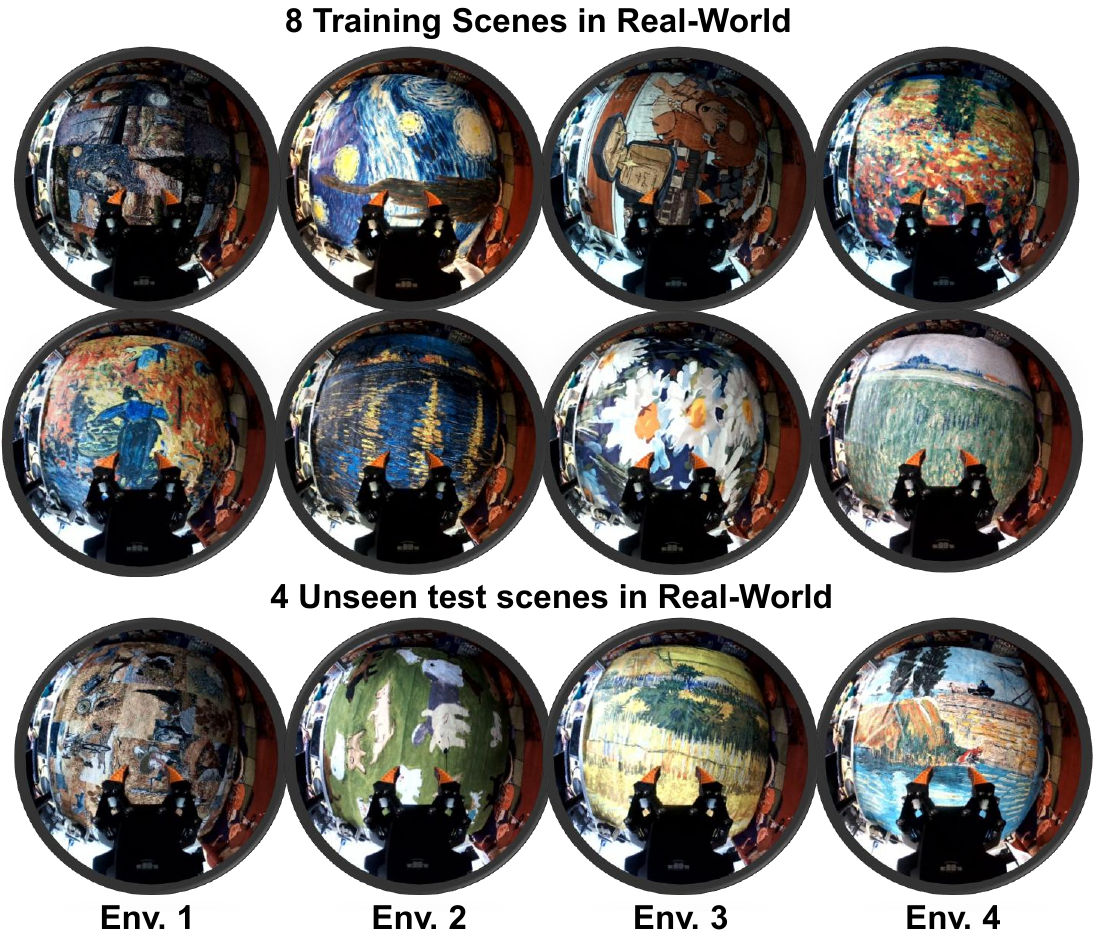}
        \vspace{0pt}
        \centerline{\small (b) Real-World Backgrounds: Patterned cloths providing diverse visual features in the real-world experiments. }
    \end{minipage}
    
    \caption{
        \textbf{Comprehensive visualization of scene diversity.} 
        We explicitly bridge the visual gap between simulation and real-world by ensuring high scene complexity in both domains. 
        \textbf{(a)} In simulation, we randomize 32 distinct textures (wood, marble, fabric, tiles) onto background surfaces. 
        \textbf{(b)} In the real world, we utilize a changeable background with visually distinct textured cloths to rigorously test spatial localization and generalization.
    }
    \label{fig:supp_visual_pall}
\end{figure*}

\textbf{Visual Setup.} 
\cref{fig:supp_visual_pall} visualizes the diverse textures employed in our study. When constructing datasets with varying numbers of scenes ($N$), A critical aspect of our Scene Generalization experiment is to isolate the benefit of environmental diversity from the benefit of increased data scale. Therefore, unlike prior work that scales up data volume~\cite{huDataScalingLaws2024}, we employ a \textit{Fixed Total Data Volume} protocol.
\begin{itemize}
    \item \textbf{Simulation Environments:} The textures were sourced from the MimicLab ~\cite{saxena2025matters} asset library for the substitution of the original scene materials.
    These range from geometric patterns to natural materials, introducing high-frequency visual features that challenge the encoder.
    A set of 32 textures was utilized for training, while a separate set of 5 previously unseen backgrounds was utilized for testing.

    We employ Coffee tasks with a fixed budget of \textbf{500 trajectories} across all experiments. As we scale the number of distinct scenes (N), we cycle through different scene renderings to ensure uniform trajectory distribution across environments. For example, in the $N=32$ setting, we employ a balanced combination of 20 scenes contributing 16 trajectories each and 12 scenes contributing 15 trajectories each, maintaining the total of 500 trajectories. 
    \item \textbf{Real-World Scenes:} We engineered a variable background system using a collection of patterned cloths (e.g., abstract art, grids) to introduce diverse visual appearances. A total of 8 distinct background scenes were constructed for training. To evaluate zero-shot generalization, we employed a separate set of 4 previously unseen background scenes for testing. 
    
     “Pick Cup” task was tested in real-word. We fix the total training budget at \textbf{200 trajectories} for all experiments. When increasing the number of unique scenes ($N$), we uniformly distribute the trajectory budget across scenes. For instance, in the $N=1$ setting, we use 200 trajectories from a single scene; in the $N=8$ setting, we use 25 trajectories from each of the 8 scenes. This ensures that any performance gain is attributable solely to the increased diversity of the visual data, rather than the quantity of demonstrations.
\end{itemize}

\begin{figure}[!ht]
    \centering
    \includegraphics[width=0.9\linewidth]{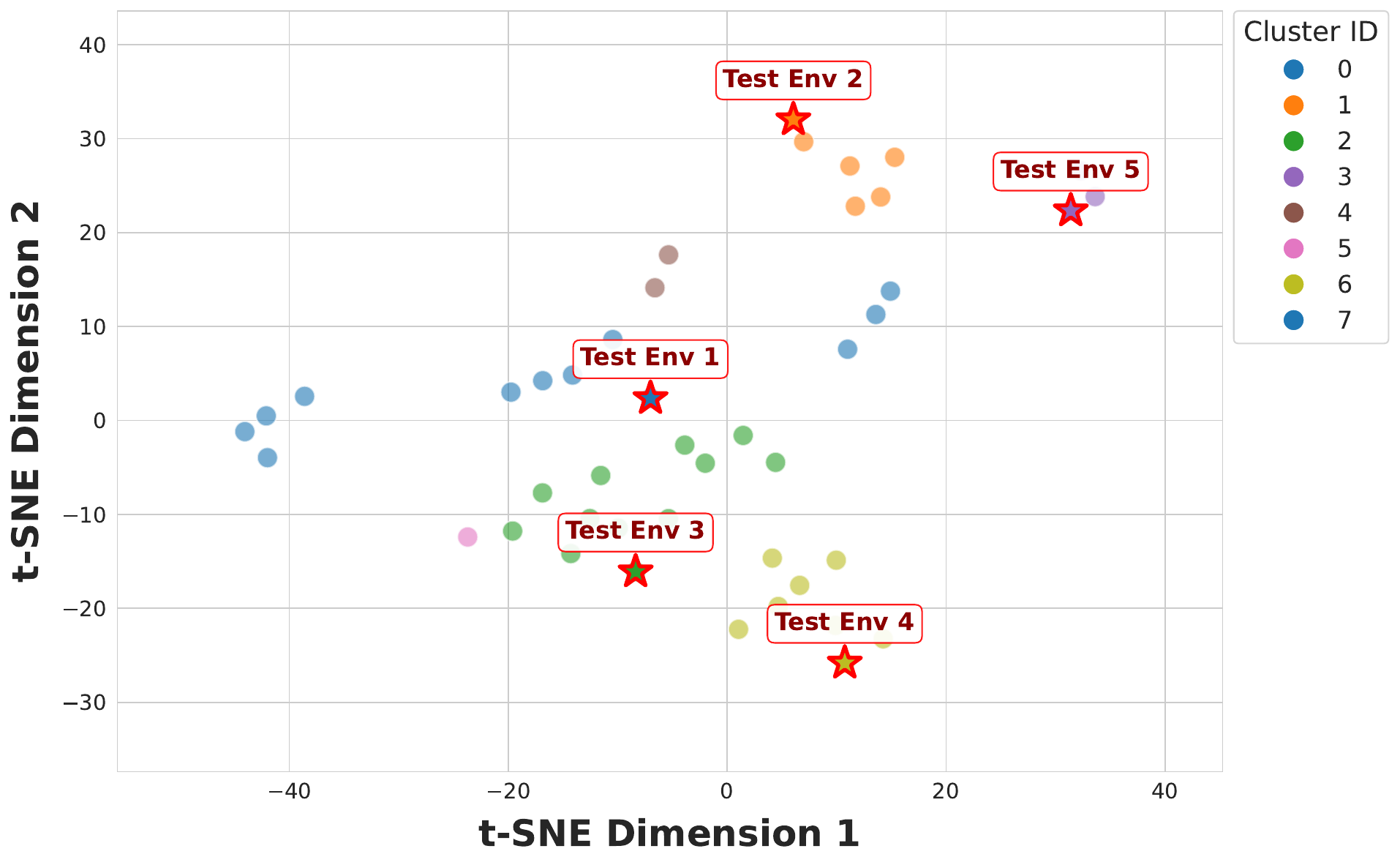} 
    \vspace{-5pt}
    \centerline{\small (a) Simulation Scene Distribution ($K=8$)}
    
    \vspace{10pt}

    \includegraphics[width=0.9\linewidth]{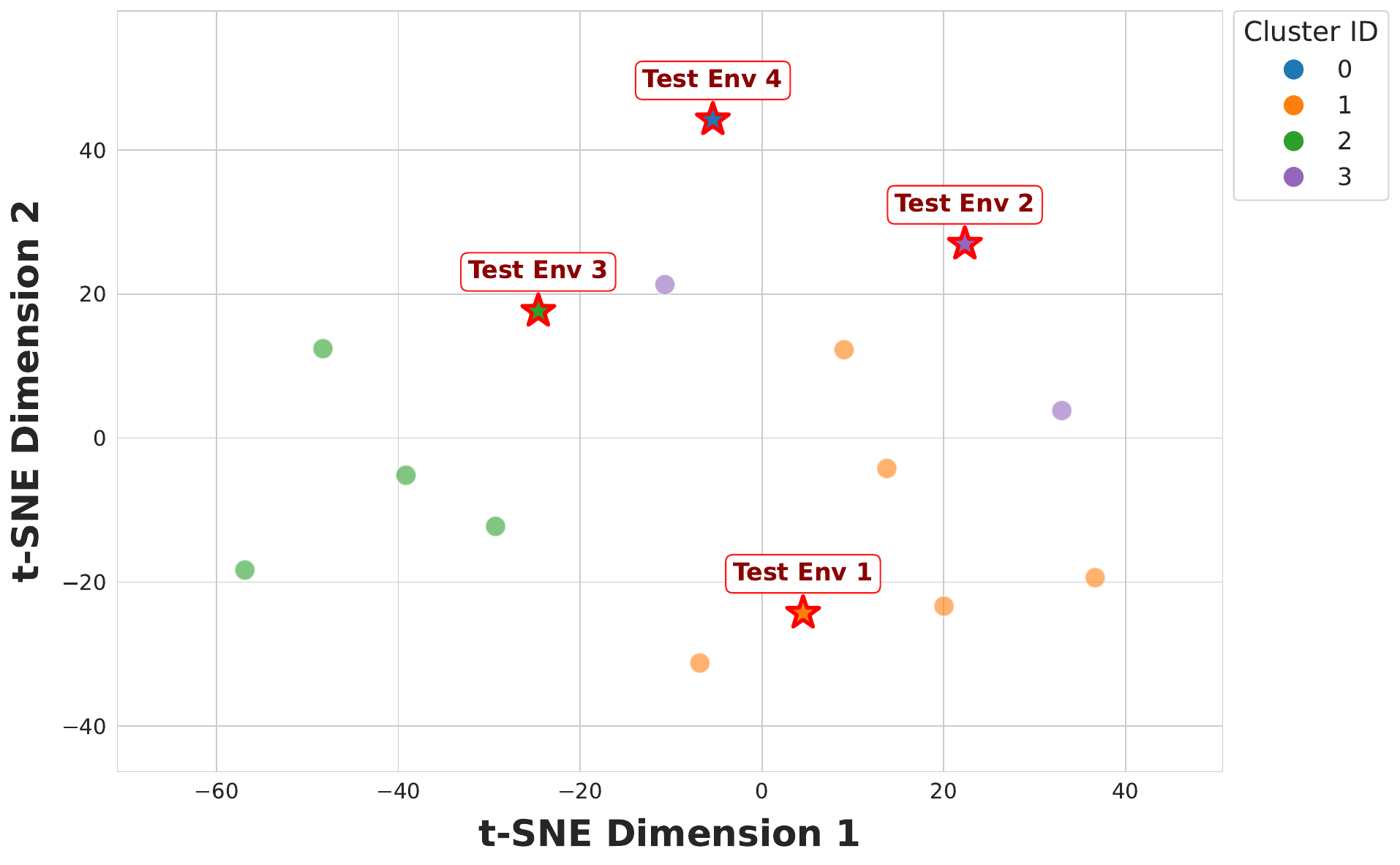} 
    \vspace{-5pt}
    \centerline{\small (b) Real-World Scene Distribution ($K=4$)}
    
  \caption{\textbf{Visualization of scene distribution in the generalization experiment.}  We visualize the scene distributions by clustering the CLIP embeddings of scene images with K-Means. The \textbf{Red Stars} highlight the \textbf{Held-out Test Scenes}.}
    \label{fig:cluster_analysis}
\end{figure}

\textbf{Data Distribution Analysis.} 
To ensure that our train/test split is statistically rigorous and covers the semantic space of possible environments, we employed a data-driven selection strategy rather than arbitrary manual selection. 
Specifically, we extracted the global semantic features of all candidate scenes using the CLS token of a pre-trained CLIP visual encoder. We then performed K-Means clustering on these embeddings to identify distinct visual clusters.
\begin{itemize}
    \item \textbf{Simulation ($K=8$):} As shown in \cref{fig:cluster_analysis}(a), the simulation textures cluster into 8 distinct groups.
    \item \textbf{Real-World ($K=4$):} As shown in \cref{fig:cluster_analysis}(b), the real-world scene cluster into 8 distinct groups.
\end{itemize}
To construct a balanced evaluation protocol, we sampled exactly one representative scene from \textit{each} cluster to serve as the \textbf{Held-out Test Set} (indicated by red boxes), while the remaining scenes formed the training set. This method guarantees that the test set is not biased towards any specific texture type and rigorously tests the policy's ability to generalize across the full spectrum of visual distributions. We adopted the settings $N \in \{1, 8, 16, 32\}$ for simulation experiments and $N \in \{1, 2, 4, 6, 8\}$ for real-world experiments, ensuring progressively diverse environmental coverage. 

\subsection{Per-Scene Generalization Analysis}

\textbf{Motivation.} 
In the main paper, we demonstrated that increasing the diversity of training scenes ($N$) significantly improves the average zero-shot generalization performance in unseen environments. However, an aggregated mean metric can potentially obscure the variance in difficulty across different test scenes. For instance, a policy might perform exceptionally well on a visually simple background while failing on a more complex one, creating a misleadingly high average. To investigate the consistency of generalization, we provide a granular, disaggregated analysis, plotting performance scaling curves for \textit{each individual} unseen test scene.

\begin{figure}[t]
    \centering

    \includegraphics[width=0.9\linewidth]{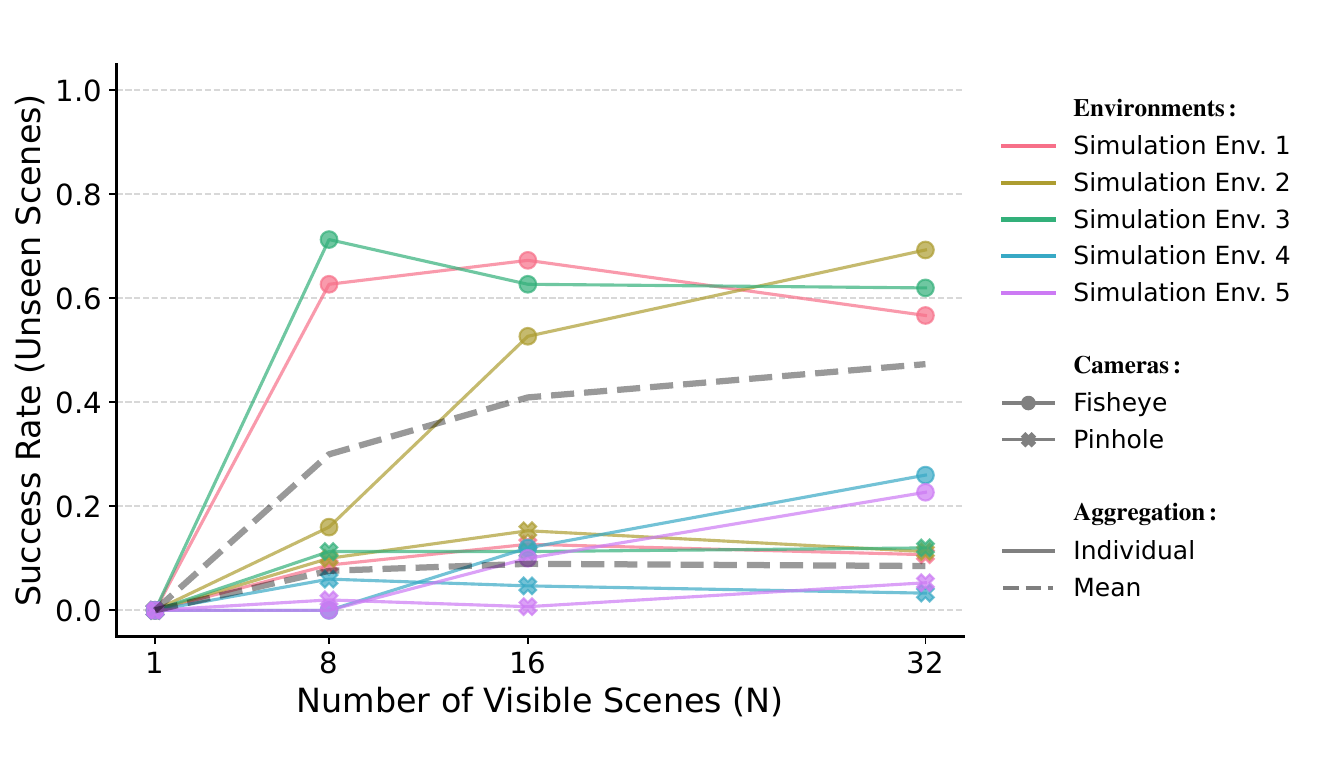} 
    \vspace{-5pt}
    \centerline{\small (a) \textbf{Simulation Task (5 Scenes)}}
    
    \vspace{15pt}

    \includegraphics[width=0.9\linewidth]{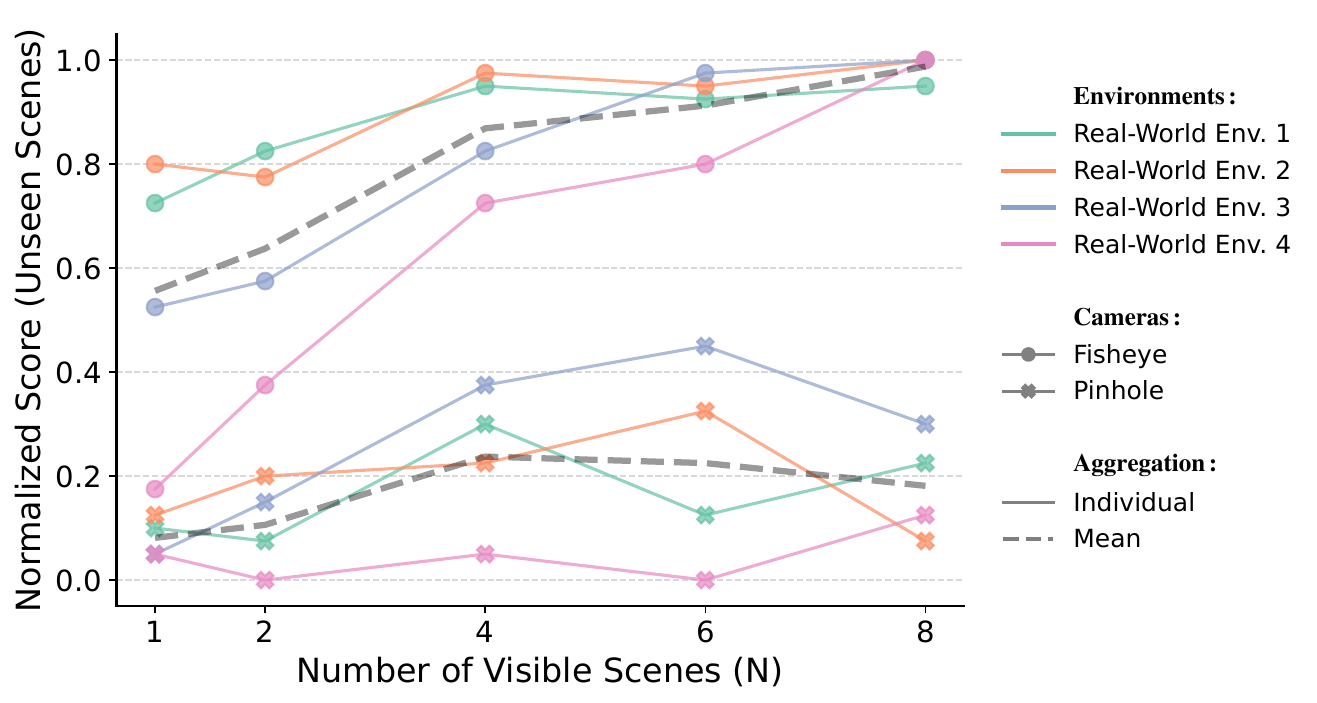} 
    \vspace{-5pt}
    \centerline{\small (b) \textbf{Real-World Task (4 Scenes)}}
    
    \vspace{5pt}

    \caption{
        \textbf{Generalization Performance on each scene.} 
        We visualize the performance scaling curves for \textit{each individual} held-out test scene to investigate the consistency of generalization.
        The \textbf{Dashed Lines} represent the aggregated mean performance, while solid lines represent specific test environments.
    }
    \label{fig:scene_breakdown}
\end{figure}

\textbf{Analysis Setup.} 
We decouple the aggregated results from Figure 8 into specific performance trajectories for every held-out environment:
\begin{itemize}
    \item \textbf{Simulation Breakdown (5 Curves):} We report the success rate scaling on each of the 5 unseen simulation test scenes (shown in \cref{fig:scene_breakdown}) as the training diversity increases ($N \in \{1, 8, 16, 32\}$).
    \item \textbf{Real-World Breakdown (4 Curves):} We report the normalized score scaling on each of the 4 unseen real-world scenes (shown in \cref{fig:scene_breakdown})—including the challenging ``Starry Night'' and ``Sunflowers'' patterns—as the training diversity increases ($N \in \{1, 2, 4, 6, 8\}$).
\end{itemize}
These per-scene visualizations aim to verify whether the fisheye camera's wide FoV confers a universal generalization advantage that is robust across distinct visual distributions, rather than being specific to certain texture types.

\subsection{Analysis of Environmental Complexity and Encoder Selection}
\label{sec:analysis-of-env}
To address the impact of environmental variations on policy performance, we provide a quantitative analysis of scene representativeness and justify the selection of visual encoders across different domains. We utilize the average ORB\cite{6126544} feature density (pts/frame) as a metric to quantify the ``visual richness" of each environment.
This metric allows us to address two critical questions: \textbf{Scene Representativeness:} By mapping success rates to specific feature densities, we evaluate whether the advantages of fisheye cameras persist in typical real-world environments rather than being confined to extreme cases (\cref{tab:typical_scenes}). \textbf{Encoder Selection:} Feature density provides a technical justification for our use of different visual encoders across domains. It reveals a vast complexity gap between simulation and real-world environments, which necessitates scaling the model's representation capacity—moving from ResNet-18 in sparse simulation to CLIP in the feature-rich real world—to effectively process the captured context (\cref{tab:encoder_ablation}).\\
\noindent
\textbf{Scene Representativeness and Texture Sensitivity:}
As summarized in \cref{tab:typical_scenes}, we evaluated the policies in real-world environments with varying degrees of visual texture. Even in ``Typical" settings, such as a standard wooden desk with median feature density ($2111.43 \pm 341.87$), the fisheye policy consistently outperforms the pinhole baseline by a significant margin ($+0.4250$). While the performance gain is most pronounced in highly textured environments ($+0.8062$), the robust success in low-texture scenes confirms that our findings generalize to practical, everyday deployment settings.

\begin{table}[h]
\centering
\scriptsize
\setlength{\tabcolsep}{2pt} 
\caption{Performance in Different Real-world Environments.} 
\label{tab:typical_scenes}
\begin{tabular}{lc|cc|c}
\toprule
\textbf{Typical Test Scene} & \textbf{Feature Density} & \multicolumn{2}{c|}{\textbf{Score}} & \textbf{Fisheye} \\
(Real-world Variations) & (ORB\cite{6126544} pts/frame) & Pinhole & \textbf{Fisheye} & \textbf{Improvement} \\ \midrule
Textureless (Poor) & \textit{1299.35 $\pm$ 176.81 (Low)} & 0.1250 & \textbf{0.5250} & \textit{+0.4000} \\
\textbf{Wooden Desk (Typical)} & \textit{2111.43 $\pm$ 341.87 (Median)} & 0.1250 & \textbf{0.5500} & \textit{+0.4250} \\
\textit{Highly Textured (Rich)} & \textit{$3574.99 \pm 102.54$(High)} & 0.1813 & \textbf{0.9875} & \textit{+0.8062} \\
\bottomrule
\end{tabular}
\end{table}
\textbf{Visual Encoder Choice:}
The choice of visual encoders (ResNet-18 for simulation and CLIP for real-world) stems from the vast discrepancy in visual complexity between domains. As quantified in \cref{tab:encoder_ablation}, real-world scenes exhibit a feature density nearly 13 times higher than our simulated environments ($3574.99$ vs. $268.77$). While a lightweight ResNet-18 suffices for processing visually sparse simulation data, the high-density information captured by fisheye lenses in the real world necessitates the robust representation capabilities of CLIP. Crucially, as shown in the ablation results, the fisheye configuration consistently outperforms the pinhole baseline across both domains, regardless of the encoder used, demonstrating that the benefits of a wide Field of View (FoV) are independent of the specific neural architecture.
\begin{table}[h]
\centering
\scriptsize
\setlength{\tabcolsep}{1pt}
\caption{Ablation Study on Visual Encoders.}
\label{tab:encoder_ablation}
\begin{tabular}{l|cc|cc}
\toprule
\textbf{Domain} & \multicolumn{2}{c|}{\textbf{Simulation (Success Rate)}} & \multicolumn{2}{c}{\textbf{Real-World (Mean Score)}} \\
\textbf{Feat. Density}(ORB\cite{6126544} pts/frame) & \multicolumn{2}{c|}{$268.77 \pm 30.27$} & \multicolumn{2}{c}{$3574.99 \pm 102.54$} \\
\midrule
\textbf{Encoder} & ResNet-18 & \textbf{CLIP} & ResNet-18 & \textbf{CLIP} \\ \midrule
Pinhole & 0.4467 & 0.5333 & 0.4250 & 0.7000 \\
Fisheye & \textbf{0.7533} & \textbf{0.77} & \textbf{0.7000} & \textbf{0.8875} \\
\bottomrule
\end{tabular}
\end{table}
\noindent
% =================================================================
% Section F: Additional Experiments for RQ3
% =================================================================

\section{Additional Experiments for RQ3 (Hardware Generalization)}
\label{sec:supp_rq3}

In this section, we provide a comprehensive analysis of cross-camera generalization across five subsections. We first detail the experimental protocols (\cref{Simulation Protocols for Cross-Camera}). To uncover the underlying causes of cross-camera failure, we perform scale sensitivity analyses in both simulation (\cref{Mechanism Analysis}) and real-world settings (\cref{real-world-cross}). Leveraging insights from these analyses, we demonstrate the effectiveness of Random Scale Augmentation in mitigating scale overfitting within a simulated environment (\cref{effectiveness-random-scale}). Finally, we present extensive zero-shot real-world verification across diverse physical camera (\cref{zero-shot cross-camera}), confirming the practical robustness of our proposed approach.

% -----------------------------------------------------------------
% F.1 Simulation Protocols
% -----------------------------------------------------------------
\subsection{Simulation Protocols for Cross-Camera Evaluation}
\label{Simulation Protocols for Cross-Camera}
\begin{figure}[htbp!]
  \centering
   \includegraphics[width=0.75\linewidth]{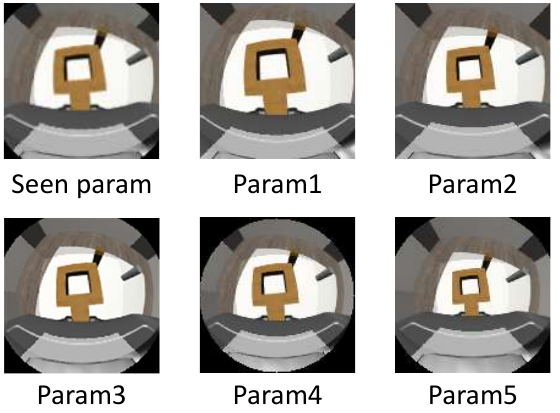}

   \caption{\textbf{Simulated fisheye effects using projection models with different distortion parameters.} Each subfigure demonstrates the visual output when applying a distinct set of parameters: ``Seen param'' represents a baseline parameter set, while ``Param1'' to ``Param5'' illustrate variations in distortion intensity, field of view, and other projection characteristics, showcasing how different parameter configurations alter the fisheye rendering.}
   \label{fig:diff_fisheye_camera}
\end{figure}

\textbf{Camera Modeling.} 
To rigorously evaluate the zero-shot hardware generalization, we established ``seen'' camera configurations for training and prepared a diverse set of ``unseen'' configurations for testing. After acquiring the equirectangular image, we can simulate various fisheye effects by applying projection models with different distortion parameters. As shown in \cref{tab:sim_camera_params}, we employ different fisheye models with configured parameters to mimic the geometric domain shifts encountered when changing lenses (e.g., switching from a wide-angle fisheye lens to a narrower one).
The \cref{fig:diff_fisheye_camera} shows the visualization effects of simulations with different parameters.

The specific models used are as follows:
\begin{itemize}
    \item \textbf{Extended Unified Camera Model (EUCM):} This model is an extension of the Unified Camera Model (UCM). By introducing two parameters, it improves the accuracy for wide field-of-view lenses. Its parameters are:
    \begin{itemize}
        \item \textbf{f:} Focal length, controlling the zoom level of the image.
        \item \textbf{a\_ :} Shape parameter with a range of (0, 1], controlling the shape of the projection curve.
        \item \textbf{b\_ :} Distortion parameter, adjusting the extent of non-linear distortion.
    \end{itemize}
    
    \item \textbf{Double Sphere Camera Model (DS):} This model describes light paths through a combination of two spheres, effectively modeling the projection geometry for large field-of-view cameras. Its parameters are:
    \begin{itemize}
        \item \textbf{f:} Focal length, controlling the zoom level of the image.
        \item \textbf{a\_ :} Blending parameter, controlling the mixing ratio between the two sphere models.
        \item \textbf{xi\_ :} Distortion parameter, representing the offset between the centers of the two spheres.
    \end{itemize}
\end{itemize}

\begin{table}[h]
\centering
\caption{\textbf{Simulation camera parameters for RQ3.} We define one seen configuration for training and distinct configurations for zero-shot evaluation. Variations in focal length and distortion simulate significant geometric shifts.}
\resizebox{\columnwidth}{!}{%
\label{tab:sim_camera_params}
\setlength{\tabcolsep}{5pt}
\renewcommand{\arraystretch}{1.1}
\begin{tabular}{ccccc}
\toprule
\textbf{Config Name} & \textbf{Method} & \textbf{Focal Length} & \textbf{Distortion} & \textbf{Scale} \\
\midrule
Seen Param & EUCM & $45$ & \begin{tabular}{@{}c@{}} $a\_: 0.4$ \\ $b\_: 2.0$ \end{tabular} & $0.9$ \\
\midrule
Param 1 & EUCM & $60$ & \begin{tabular}{@{}c@{}} $a\_: 0.5$ \\ $b\_: 2.0$ \end{tabular} & $1.0$ \\
\midrule
Param 2 & DS & $50$ & \begin{tabular}{@{}c@{}} $a\_: 0.5$ \\ $xi\_: 0.1$ \end{tabular} & $1.0$ \\
\midrule
Param 3 & EUCM & $45$ & \begin{tabular}{@{}c@{}} $a\_: 0.4$ \\ $b\_: 2.0$ \end{tabular} & $1.0$ \\
\midrule
Param 4 & EUCM & $45$ & \begin{tabular}{@{}c@{}} $a\_: 0.4$ \\ $b\_: 2.5$ \end{tabular} & $1.0$ \\
\midrule
Param 5 & EUCM & $35$ & \begin{tabular}{@{}c@{}} $a\_: 0.4$ \\ $b\_: 1.2$ \end{tabular} & $1.0$ \\
\bottomrule
\end{tabular}%
}
\end{table}

% -----------------------------------------------------------------
% F.2 Mechanism Analysis
% -----------------------------------------------------------------
\subsection{Mechanism Analysis: Scale Overfitting vs. Scale Invariance}
\label{Mechanism Analysis}
\textbf{Hypothesis.} In the main paper, we hypothesized that the primary failure mode for cross-camera transfer is \textbf{``Scale Overfitting.''} Since different lens intrinsics project objects at different sizes (e.g., a wider FoV makes objects appear smaller), a standard policy overfits to the absolute pixel scale of objects in the training set. When the lens changes, this absolute scale prior breaks, leading to failure.

\begin{figure}[h]
    \centering
    \includegraphics[width=\linewidth]{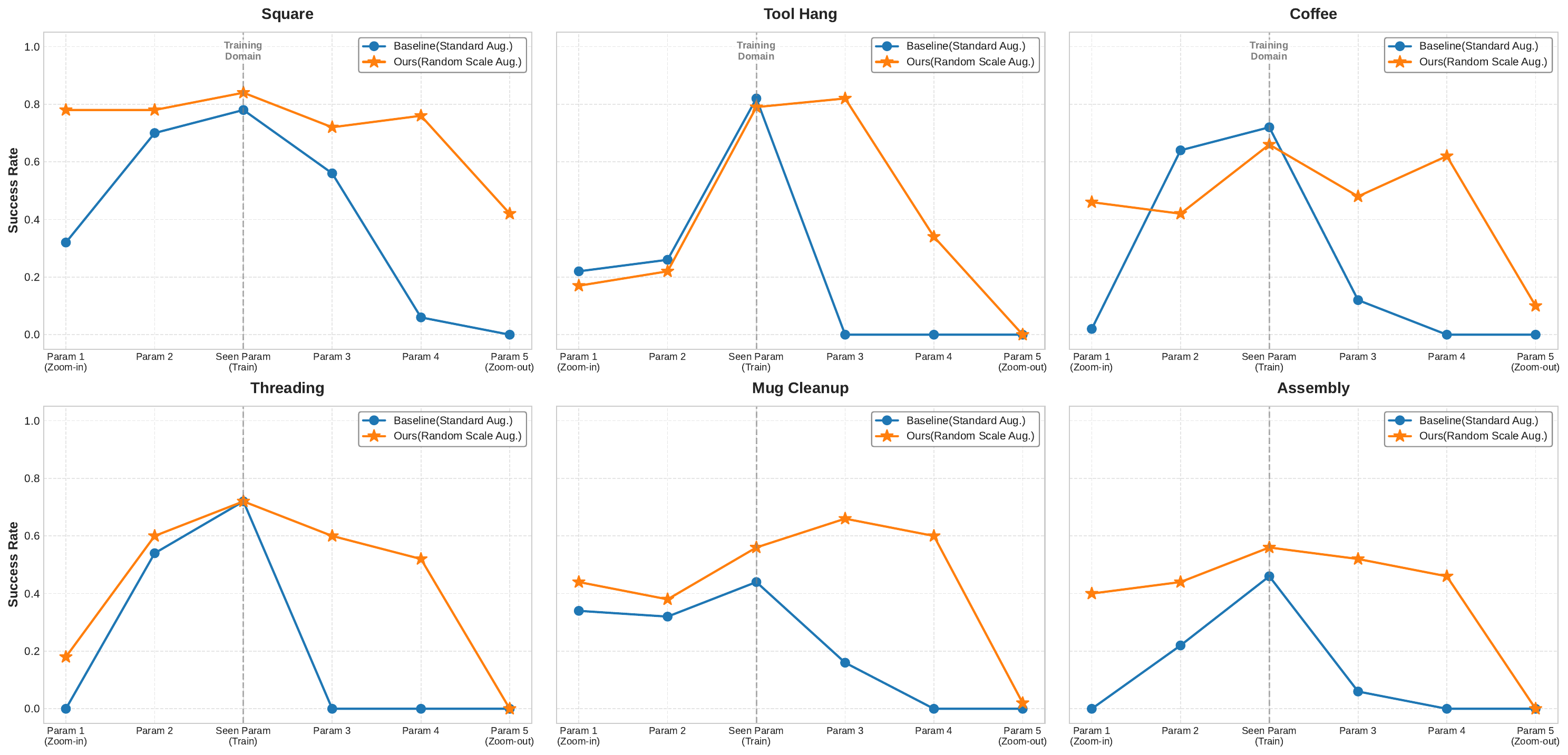}
    \caption{\textbf{Cross-Camera generalization performance in simulation for each task.} We evaluate the zero-shot camera-transfer performance of policies trained on a single ``Seen Param'' configuration across several unseen camera settings. The \textbf{Baseline (Blue)} exhibits a sharp performance drop as camera parameters deviate from the training domain, indicating severe overfitting to absolute object scales. In contrast, our \textbf{RSA method (Orange)} demonstrates a broad generalization plateau, maintaining robust performance across a wide range of unseen camera parameters.}
    \label{fig:scale_sensitivity}
\end{figure}

\textbf{Quantitative Analysis.} 
To verify this, we visualize the performances for all six tasks in simulation as shown in \cref{fig:scale_sensitivity}.

\begin{itemize}
    \item \textbf{Baseline Failure (The ``Peaked'' Curve):} The policy trained with standard augmentation (Blue Line) performs well \textit{only} on the ``Seen Param'' and ``Param 2'', which has a similar visual scale. Performance collapses on configurations that introduce significant scale shifts (e.g., Param 4 and Param 5), confirming that the policy relies heavily on specific geometric cues from the training lens.
    \item \textbf{RSA Success (The ``Plateau'' Curve):} In contrast, the policy trained with our \textbf{Random Scale Augmentation (RSA)} (Orange Line) maintains a high success rate across a wide range of parameters. RSA effectively forces the network to learn \textbf{scale-invariant features} (e.g., relative spatial relationships) rather than memorizing absolute pixel sizes.
\end{itemize}

% -----------------------------------------------------------------
% F.3 Real-World Verification
% -----------------------------------------------------------------
\subsection{Real-World Cross-Camera Verification}
\label{real-world-cross}

To verify our hypothesis in simulation, we conducted a scale sensitivity analysis on the real robot. Since physically swapping sensor modules to precisely control intrinsic parameters is impractical, we simulate geometric domain shifts by applying varying center-crop scale factors ($S$) to the input images during inference, effectively mimicking changes in focal length and FoV. This operation effectively mimics the ``Zoom In'' (narrower FoV) and ``Zoom Out'' (wider FoV) effects caused by changing lens parameters. We evaluated the policy's performance across a scale range of $S \in [0.7, 1.3]$, where $S=1.0$ represents the training distribution.
\begin{figure}[t]
    \centering
\includegraphics[width=\linewidth]{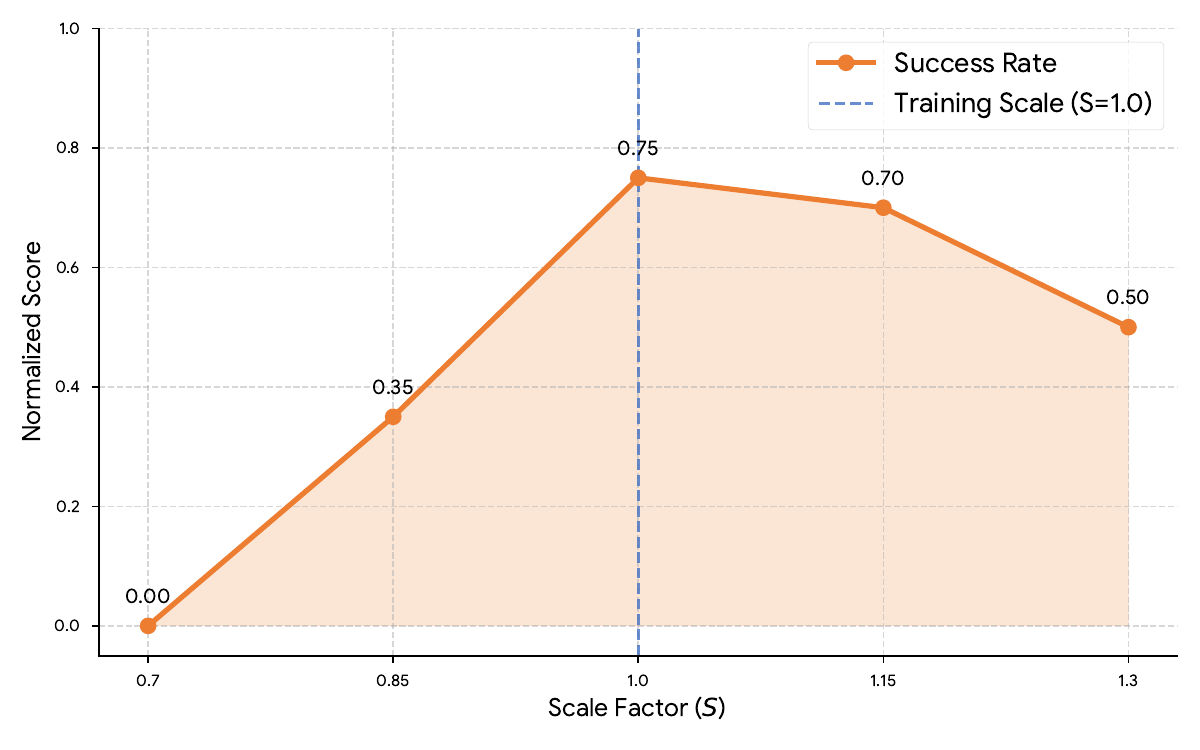}
    
    \vspace{-5pt}
    
    \caption{
        \textbf{Real-world Scale Sensitivity Analysis.} 
        To investigate the root cause of cross-camera transfer failure, we simulate geometric domain shifts by applying center crops with varying scale factors ($S$) to the fisheye input. 
        The results show a characteristic \textbf{``inverted-V'' performance drop}: the policy performs robustly near the training scale ($S=1.0$) but suffers catastrophic degradation as the scale deviates significantly (e.g., $0\%$ success rate at $S=0.7$). 
        This confirms that the standard policy is highly sensitive to absolute object scale, validating the necessity of our Random Scale Augmentation (RSA) strategy.
    }
    \label{fig:real_world_scale_sensitivity}
\end{figure}

\begin{figure}[t]
    \centering
    \includegraphics[width=1.0\linewidth]{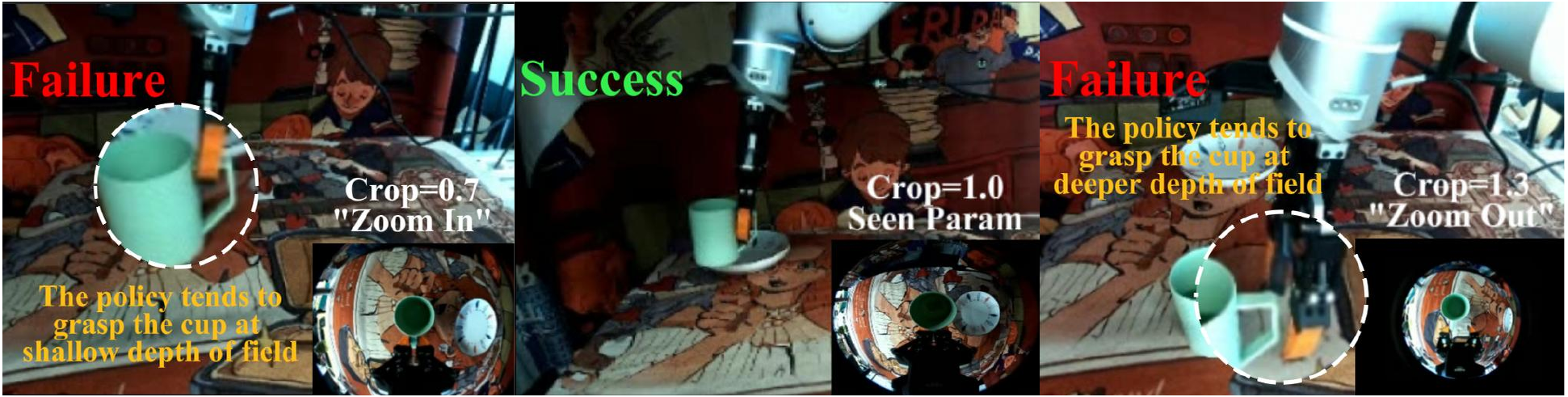}
    
    \vspace{-5pt}
    
    \caption{
        \textbf{Qualitative Visualization of Scale-Induced Failures.} 
        We visualize the specific failure modes of the baseline policy under geometric scale shifts (simulated via center cropping).
        \textbf{(Left)} When zoomed in (Crop=$0.7$), the object appears larger, misleading the policy to perceive it as closer; consequently, the robot attempts to grasp at a \textit{shallower} depth (undershooting the target).
        \textbf{(Right)} When zoomed out (Crop=$1.3$), the object appears smaller, causing the policy to perceive it as farther away; this leads to grasping at a \textit{deeper} depth (overshooting or colliding with the target).
        \textbf{(Middle)} Accurate manipulation is only achieved at the training scale (Crop=$1.0$).
        These behaviors explicitly confirm that the cross-camera failure stems from the policy's overfitting to the absolute pixel scale of objects. \textbf{Please see the video in the supplementary files for more details.}
    }
    \label{fig:rq3_qualitative_failure}
\end{figure}

\noindent \textbf{Quantitative Analysis: Sensitivity to Scale.} 
As illustrated in \cref{fig:real_world_scale_sensitivity} (Scale Sensitivity Analysis), the baseline policy exhibits a characteristic \textbf{``inverted-V'' performance curve}. While the policy maintains robust performance near the training scale ($S=1.0$, score 0.75), it suffers catastrophic degradation as the scale deviates. Notably, a ``Zoom In'' operation ($S=0.7$) causes the success rate to plummet to \textbf{0.0}, while a ``Zoom Out'' ($S=1.3$) drops the performance to \textbf{0.5}. This sharp decline confirms that the standard fisheye policy is highly sensitive to absolute object scale, corroborating our simulation hypothesis that scale overfitting is the primary bottleneck for cross-camera transfer.\\
\textbf{Qualitative Analysis: The Depth-Scale Ambiguity.} 
To understand the failure mechanism, we visualize specific rollout behaviors in \cref{fig:rq3_qualitative_failure}. The results reveal a distinct correlation between visual scale and depth estimation errors:
\begin{itemize}
    \item \textbf{Underestimation of Depth (Zoom In):} At $S=0.7$, the object appears significantly larger in the image frame. The policy misinterprets this visual cue as the object being \textit{closer} than it actually is, resulting in a grasp attempt at a \textbf{shallower depth} (undershooting the target).
    \item \textbf{Overestimation of Depth (Zoom Out):} Conversely, at $S=1.3$, the object appears smaller. The policy perceives it as being \textit{farther away}, leading to a grasp at a \textbf{deeper depth} (often colliding with or overshooting the target).
\end{itemize}
These findings provide empirical evidence that, in the absence of scale-invariant training (e.g., RSA), fisheye-based policies rely heavily on absolute pixel size for spatial reasoning, making direct cross-camera transfer inherently difficult.

\subsection{Effectiveness of Random Scale Augmentation}
\label{effectiveness-random-scale}
Building upon the scale-shift protocols defined in \cref{real-world-cross}, we evaluate the effectiveness of Random Scale Augmentation (RSA) in enhancing policy robustness. Our results demonstrate that RSA provides consistent generalization improvements, serving as a defense against geometric domain shifts that typically cause standard imitation learning policies to fail.
The detailed results of our evaluation are presented in \cref{tab:scale_sensitivity}, which quantifies the performance of different policies under the varying scale factors.
\begin{itemize}
\item \textbf{Consistent Generalization Improvements}: Policies trained with RSA demonstrate steady performance gains over the Standard Aug. (Standard Augmentation) baseline across all evaluated scales. For instance, at a scale factor of $S=0.85$, the RSA-trained Diffusion Policy achieves a score of $0.950$, whereas Standard Aug. yields $0.350$.
\item \textbf{Reducing Scale-Induced Performance Degradation}: Standard Augmentation shows sensitivity to scale shifts, particularly under zoom-in ($S=0.70$), where the Diffusion Policy's performance reaches zero. In contrast, RSA maintains a score of $0.725$ under the same conditions, mitigating failures caused by object magnification.
\item \textbf{Integration with Large-Scale Models}: When integrated with the $\pi_{0.5}$ architecture, RSA facilitates improved scale invariance. Notably, $\pi_{0.5}$ equipped with RSA maintains a score of $1.000$ even under significant zoom-out ($S=1.30$), a scenario where the Standard Aug. score falls to $0.150$.
\end{itemize}
These results suggest that RSA encourages the visual encoder to prioritize relative spatial relationships—such as the target object's size relative to the gripper—over absolute pixel scales. This prioritization supports more robust cross-hardware deployment by reducing the impact of lens-specific geometric shifts.

\begin{table}[htbp]
\centering
\caption{Normalized scores for the ``Pick and Place'' task under simulated scale shifts. We compare \textbf{Standard Aug.} (Standard Augmentation) and \textbf{RSA} (Random Scale Augmentation) across different parameters ($S$).}
\resizebox{\columnwidth}{!}{
\label{tab:scale_sensitivity}
\begin{tabular}{llccccc}
\toprule
\textbf{Policy Model} & \textbf{Aug. Strategy} & \textbf{$S=0.70$} & \textbf{Param 1} & \textbf{$S=1.0$} & \textbf{$S=1.15$} & \textbf{$S=1.30$} \\
& & (Zoom-in) & & (Seen) & & (Zoom-out) \\
\midrule
Diffusion Policy \cite{chi2024diffusionpolicy} & Standard Aug. & 0.000 & 0.350 & 0.750 & 0.750 & 0.500 \\
& \textbf{RSA (Ours)} & \textbf{0.725} & \textbf{0.950} & \textbf{1.000} & \textbf{0.750} & \textbf{0.650} \\
\midrule
$\pi_{0.5}$ \cite{black2024pi_0} & Standard Aug. & 0.375 & 0.875 & 1.000 & 0.600 & 0.150 \\
& \textbf{RSA (Ours)} & \textbf{0.900} & \textbf{1.000} & \textbf{1.000} & \textbf{0.975} & \textbf{1.000} \\
\bottomrule
\end{tabular}}
\end{table}

\subsection{Zero-shot Cross-Camera Validation}
\label{zero-shot cross-camera}
To evaluate the practical utility of RSA, we conducted zero-shot transfer experiments between different physical fisheye lenses in the real world. As detailed in \cref{tab:real_hardware}, a policy trained on a standard $180^\circ$ lens was deployed directly onto hardware equipped with Narrow ($150^\circ$) and Wide ($220^\circ$) lenses, introducing distinct geometric and scale shifts. To ensure the findings are representative of modern generalist policies, we conduct this validation using the state-of-the-art $\pi_{0.5}$ architecture.

\begin{itemize}
\item \textbf{Addressing Hardware-Induced Scale Shifts:} The Standard Aug. (Baseline) policy shows an observable performance decline when encountering hardware variations. Specifically, when transitioning to a Wide Lens ($\sim 0.8\times$ scale shift), the baseline score reaches $0.0025$. 

\item \textbf{Real-world Generalization:} In contrast, RSA mitigates these hardware-induced shifts across different physical lenses. For the Narrow Lens ($\sim 1.2\times$ scale shift), RSA increases the score from $0.5000$ to $0.9500$. Even in the more challenging Wide Lens scenario, RSA recovers the performance to $0.6000$.

\item \textbf{Practical Deployment Implications:} These results suggest that RSA is an effective strategy for real-world robotics beyond simulation-based heuristics. By decoupling the policy from absolute pixel scales, RSA facilitates the reuse of existing datasets across diverse camera systems.
\end{itemize}

\begin{table}[h]
\centering
\scriptsize
\setlength{\tabcolsep}{1pt}
\caption{Real-world Cross-camera Generalization with RSA.} 
\label{tab:real_hardware}
\begin{tabular}{lcccc}
\toprule
\textbf{Test Camera} & \textbf{FOV Angle} & \textbf{Induced Scale Shift} & \textbf{Baseline(Standard Aug.)} & \textbf{RSA (Ours)} \\ \midrule
Seen Camera & 180$^\circ$ & 1.0$\times$ (Seen) & 1.0000 & \textbf{1.0000} \\
Narrow Lens & 150$^\circ$ & $\sim$1.2$\times$ (Zoom In) & 0.5000 & \textbf{0.9500} \\
Wide Lens & 220$^\circ$ & $\sim$0.8$\times$ (Zoom Out) & 0.0025 & \textbf{0.6000} \\ \bottomrule
\end{tabular}
\end{table}
In summary, these findings demonstrate that RSA effectively bridges the gap between diverse hardware configurations. By encouraging the learning of scale-invariant features, RSA offers a practical path for deploying vision-based policies across varied robotic platforms without the need for hardware-specific fine-tuning.

\end{document}